\documentclass[letterpaper]{article} 
\usepackage{aaai2026}  
\usepackage{times}  
\usepackage{helvet}  
\usepackage{courier}  
\usepackage[hyphens]{url}  
\usepackage{graphicx} 
\usepackage{natbib}  
\usepackage{caption} 
\usepackage{algorithm}
\usepackage{algorithmic}
\usepackage{amsmath}
\usepackage{booktabs}
\usepackage{subcaption}
\usepackage{siunitx}
\usepackage{comment}
\usepackage{multirow}
\usepackage{amsthm}
\usepackage{amssymb}
\usepackage[utf8]{inputenc}
\DeclareUnicodeCharacter{21D2}{\ensuremath{\Rightarrow}}

\nocopyright

\theoremstyle{definition}
\newtheorem{theorem}{Theorem}

\usepackage{newfloat}
\usepackage{listings}
\DeclareCaptionStyle{ruled}{labelfont=normalfont,labelsep=colon,strut=off} 
\floatstyle{ruled}
\newfloat{listing}{tb}{lst}{}
\floatname{listing}{Listing}

\title{CSP4SDG: Constraint and Information-Theory Based Role Identification in Social Deduction Games with LLM-Enhanced Inference}
\author{
    Kaijie Xu\textsuperscript{\rm 1},
    Fandi Meng\textsuperscript{\rm 2},
    Clark Verbrugge\textsuperscript{\rm 1},
    Simon Lucas\textsuperscript{\rm 2}\\
}
\affiliations{
    \textsuperscript{\rm 1}Department of Computer Science, McGill University, Montreal, Quebec, Canada\\
    \textsuperscript{\rm 2}Queen Mary University of London, London, United Kingdom\\
    kaijie.xu2@mail.mcgill.ca, f.meng@qmul.ac.uk, clump@cs.mcgill.ca, simon.lucas@qmul.ac.uk
}

\begin{document}

\maketitle

\begin{abstract}
In Social Deduction Games (SDGs) such as \emph{Avalon}, \emph{Mafia}, and \emph{Werewolf}, players conceal their identities and deliberately mislead others, making hidden-role inference a central and demanding task. Accurate role identification, which forms the basis of an agent's belief state, is therefore the keystone for both human and AI performance. We introduce \textbf{CSP4SDG}, a probabilistic, constraint–satisfaction framework that analyses gameplay objectively. Game events and dialogue are mapped to four linguistically-agnostic constraint classes—\emph{evidence}, \emph{phenomena}, \emph{assertions}, and \emph{hypotheses}. Hard constraints prune impossible role assignments, while weighted soft constraints score the remainder; information-gain weighting links each hypothesis to its expected value under entropy reduction, and a simple closed-form scoring rule guarantees that truthful assertions converge to classical hard logic with minimum error. The resulting posterior over roles is fully interpretable and updates in real time.
Experiments on three public datasets show that CSP4SDG (i) outperforms LLM-based baselines in every inference scenario, and (ii) boosts LLMs when supplied as an auxiliary ``reasoning tool.''
Our study validates that principled probabilistic reasoning with information theory is a scalable alternative—or complement—to heavy-weight neural models for SDGs. 
\end{abstract}

\begin{links}
    \link{Code}{https://github.com/Nortrom1213/CSP4SDG}
\end{links}

\section{Introduction}

Social Deduction Games (SDGs), like \textit{Avalon}, \textit{Mafia}, and \textit{Werewolf}, require players to infer hidden roles despite deception and sparse evidence. Although recent efforts cover transformer role classifiers, reinforcement learning agents, and planning‑LLM hybrids
\cite{de2018mafiascum,serrino2019finding,ibraheem2022putting,lai2023werewolf}, they typically treat chat as opaque text, require heavy training, or embed game‑specific heuristics—hindering interpretability and cross‑title transfer.  

We propose \textbf{CSP4SDG}, a training‑free, probabilistic constraint‑satisfaction framework.  A lightweight LLM converts raw logs into four language‑agnostic constraint types; hard constraints prune impossible worlds, and soft ones receive information‑gain weights that remove manual tuning required by valued or VOI-CSPs \cite{schiex1995valued,mackay1992information}.  The solver returns calibrated posteriors and MAP assignments that stand alone or refine the LLM, yielding an interpretable, game-agnostic, plug-and-play module.

To test the generality of our approach, we run a unified evaluation on three public SDG datasets that cover crowd-sourced chat and large-scale log files.  On each dataset, we compare three reasoning methods—\emph{pure CSP}, \emph{LLM-only}, and the hybrid \emph{LLM+CSP}—while ablating the CSP solver through five settings.  Every experiment is repeated under multiple player perspectives (objective, good-roles, evil-roles) and both truthful- and deceptive-good conditions.  Finally, we perform a backbone ablation with successively stronger LLMs, isolating  the effect of language-model capacity.  This comprehensive design allows us to examine (i) how much structure alone can achieve under different settings, (ii) how much an LLM can learn without structure, and (iii) how the two components interact when combined. 

\textbf{Our main contributions are as follows:}
\begin{itemize}
    \item \textbf{Generalized Probabilistic CSP Framework for Role Inference:} We formulate SDG role inference as a training-free probabilistic constraint-satisfaction problem that integrates logical filtering and information-theoretically weighting in a single, generic framework.   

    \item \textbf{LLM-driven end-to-end workflow:} We design a lightweight LLM pipeline that converts raw game logs into structured constraints and seamlessly couples them with the CSP solver, yielding an interpretable, plug-and-play reasoning module.  

    \item \textbf{Empirical Validation Across Multiple Datasets:} Experiments on three public datasets—covering various CSP settings, diverse viewpoints, and several LLM backbones—show that CSP4SDG consistently outperforms baseline methods and reliably boosts LLM reasoning.
\end{itemize}
\begin{figure*}[t]
  \centering
  \includegraphics[width=\linewidth]{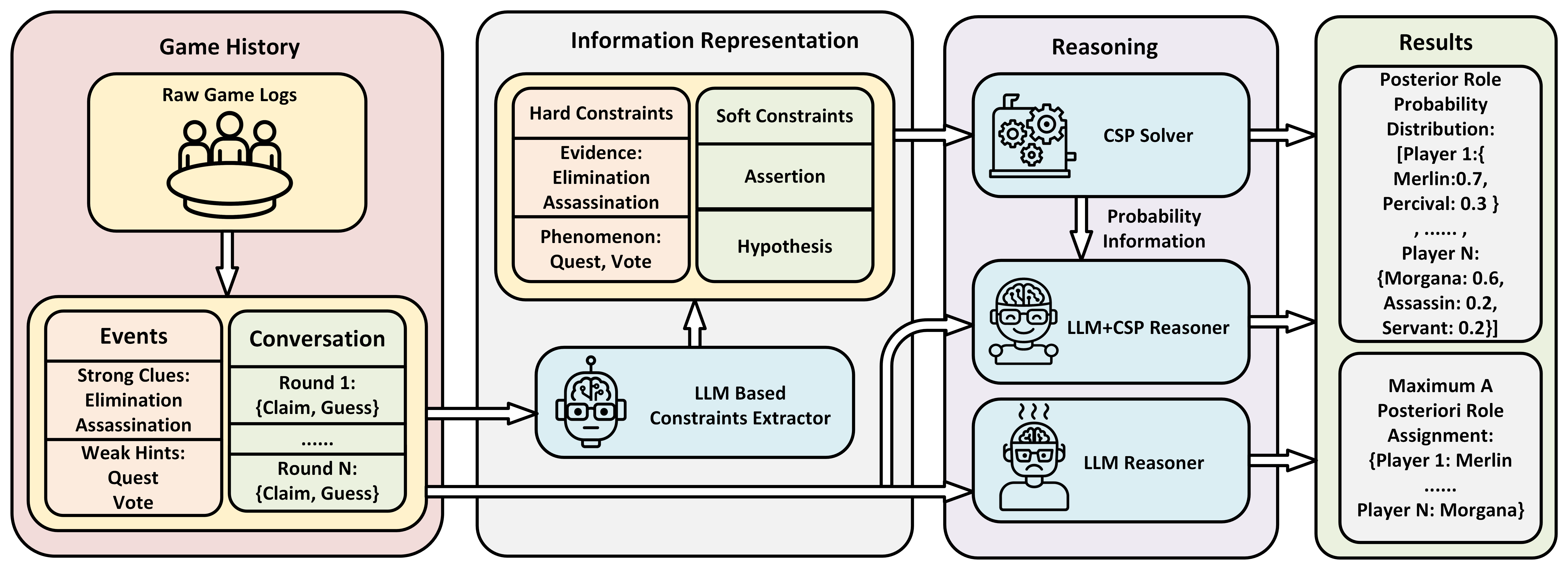}
\caption{%
Schematic overview of the proposed inference architecture for social–deduction games.  
\textbf{Game History} (left) comprises objective \emph{event} traces (e.g., eliminations, assassinations, quest outcomes) and subjective \emph{conversation} turns.  
\textbf{Information Representation} (centre) leverages an auxiliary LLM \cite{mann2020language} to transform raw logs into a structured constraint set:  
(i) \emph{hard constraints} (evidence and phenomena) that are logically inviolable, and  
(ii) \emph{soft constraints} (player assertions and hypotheses) that contribute graded probabilistic weight.  
\textbf{Reasoning} (right) contrasts three inference engines.  
A \emph{plain LLM} lacks the combinatorial apparatus required for reliable role deduction;  
a \emph{hybrid LLM\,+\,CSP} benefits from externally supplied posteriors but is still bottlenecked by heuristic language reasoning;  
our \emph{CSP solver} enforces all hard constraints and optimally scores soft ones, delivering calibrated posterior distributions and MAP role assignments that achieve the highest empirical accuracy.}
  \label{fig:pipeline-overview}
\end{figure*}
\section{Related Works}

Extensive research has been conducted in SDGs.  Early studies extracted lexical and pragmatic deception cues \cite{zhou2008cues,niculae2015linguistic}; the Mafiascum corpus enabled supervised classifiers \cite{de2018mafiascum}, later extended to transformer and LLM benchmarks \cite{ibraheem2022putting,stepputtis2023long}.  Multimodal persuasion \cite{lai2023werewolf}, reinforcement‑learning from logs \cite{serrino2019finding}, mental‑state agents \cite{nakamura2016constructing}, endgame SVMs \cite{chuchro2022training}, and planning–LLM hybrids \cite{meta2022human} further diversified the field; Other works explore SDG research in multiple dimensions \cite{kopparapu2022hidden,eger2018keeping,chi2024amongagents,sarkar2025training,kim2024fine,velikov2021rlerewolf,martinenghi2024llms,wu2024enhance,carminati2023hidden}.  Yet most approaches rely on heavy task‑specific training, treat dialogue as opaque text, or lack principled uncertainty—gaps our CSP4SDG fills with a training‑free, interpretable, and game‑agnostic reasoning module. A complete survey appears in \textbf{Appendix A}.

Classical constraint satisfaction offers exact, interpretable search \cite{freuder2006constraint}; soft \cite{schiex1995valued} and probabilistic \cite{fargier1996mixed} variants rank near‑consistent worlds but depend on hand‑tuned costs.  Information‑theoretic criteria quantify evidential value \cite{mackay1992information}.  CSP4SDG unifies these strands: LLM‑extracted constraints become IG‑weighted soft relations, eliminating manual costs and yielding calibrated posteriors without retraining.

\section{Methods}

\subsection{Problem Definition and Framework}

SDGs involve a group of players secretly assigned different roles. Each player aims to deduce the hidden roles of others through various events and conversations. Precise \emph{role-level} inference—not merely a good/evil split—is pivotal in SDGs with multiple special abilities: knowing a player’s role lets teammates interpret their actions as structured evidence and coordinate strategy, while adversaries can exploit that same knowledge to craft targeted deceptions that mislead opponents who possess only partial information. To facilitate and automate the role identification process, we propose a generalized constraint-based framework for various SDGs.

\subsubsection{General Formulation}

An SDG scenario is defined by:
\begin{itemize}
    \item A set of players: \( P = \{p_1, p_2, \dots, p_n\} \).
    \item A set of roles: \( R = \{r_1, r_2, \dots, r_m\} \), partitioned into ``good'' (e.g., Loyal Servant, Villager) and ``evil'' roles (e.g., Assassin, Mafia).
    \item A world-state assignment: \( a: P \rightarrow R \), which assigns each player a role from the role set \( R \).
\end{itemize}

Given observed game events and players' utterances at time \(t\), our task is to infer and update the posterior distribution of players' roles
$Pr(r \mid \mathcal{C}_t) $ for each player-role pair $(p_i, r)$, where \(\mathcal{C}_t\) represents all observed constraints until time \(t\). This role identification task can serve two purposes:
\begin{enumerate}
    \item \textbf{Direct Role Prediction}: To directly infer the most probable roles of each player to support automated agents in making in-game decisions or to analyze player strategies in post-game scenarios.
    \item \textbf{Auxiliary Decision Support}: To provide probabilistic insights as auxiliary input for human players or for models (such as LLMs) to enhance their reasoning capability and potential decision-making accuracy during the game.
\end{enumerate}

\subsubsection{Constraint-Based Representation}

Our proposed generalized framework categorizes in-game information into four constraint types. Each constraint type has a clear semantic interpretation, logical formulation, and standardized template, as illustrated in Table~\ref{table:constraint_summary}.

The distinction between these constraint types lies primarily in their certainty level and consequently, their treatment in the inference process:
\begin{itemize}
    \item \textbf{Evidence (E) and Phenomenon (P)}: Evidence fixes a player’s role exactly, whereas Phenomenon narrows each player’s role domain; both are hard constraints that prune any assignment violating them. 
    \item \textbf{Assertions (A)}: These player-made statements are treated as soft constraints with extremely high weights (\(w_A\gg1\)). Assignments satisfying more assertions are exponentially favored, ensuring that satisfying assertions is highly prioritized to hypotheses.
    \item \textbf{Hypotheses (H)}: Representing weaker, player-driven speculations or supports; assigned relatively low weights. They provide subtle probabilistic preferences to the inference without strictly excluding possibilities.
\end{itemize}

\begin{table*}[htbp]
\centering\small
\begin{tabular}{|l|l|l|l|l|}
\hline
\textbf{Type} & \textbf{Game} & \textbf{Event / Dialogue Cue} & \textbf{Constraint Format (grammar)} & \textbf{Mathematical Representation}\\
\hline
Evidence & Avalon & Assassin kills Merlin & \textit{role\_is(p1/p2,assassin/merlin)} & 
$a\models\bigl(\text{role}(p_1, p_2)=[\text{assassin}, \text{merlin}]\bigr)$\\
\cline{2-5}
 & Mafia  & Night victim revealed & \textit{role\_is(p, Bystander)} & $a\models\text{role}(p)=\text{Bystander}$\\
 \cline{2-5}
\hline
Phenomenon & Avalon & Quest $q$ has $f$ fail cards & \textit{evil\_at\_least(team,\,f)} & $a\models \sum_{p\in \text{team}}\mathbf 1[\text{evil}(p)]\ge f$\\
\hline
Assertion & Avalon & “I am Percival.” & \textit{assert\_role\_is(speaker, role)} & $S(a)\,\times =\,w_A\;\mathbf 1[a\models\text{role}(s)=\text{Percival}]$\\
\cline{2-5}
 & Avalon & “This team is clean.” & \textit{assert\_team\_good(speaker, team)} & $S(a)\,\times =\,w_A\;\mathbf 1[\forall p\in\text{team}\,\text{good}(p)]$\\
\cline{2-5}
 & Avalon & “X is evil.” & \textit{assert\_role\_in(sp,\,tg,\,evil)} & $S(a)\,\times =\,w_A\;\mathbf 1[\text{evil}(tg)]$\\
\hline
Hypothesis  & Avalon & Weak suspicion / NO vote & \textit{hypo\_role\_in(sp,\,tg,\,evil)} & $S(a)\;+\!=w_{mid}\,\mathbf 1[\text{evil}(tg)]$\\
\cline{2-5}
 & Avalon & Weak support / YES vote & \textit{hypo\_role\_in(sp,tg,good)} & $S(a)+=w_{low}\,\mathbf 1[\text{good}(tg)]$\\
\cline{2-5}
 & Avalon & Proposer chooses squad & \textit{hypo\_team\_good(proposer, team)} & $S(a)+=w_{strong}\,\mathbf 1[\forall p\in\text{team}\,\text{good}(p)]$\\
 \cline{2-5}
 & Mafia  & “X looks Mafia.” & \textit{hypo\_role\_in(sp,\,tg,\,mafia)} & $S(a)+=w_{mid}\,\mathbf 1[\text{mafia}(tg)]$\\
\cline{2-5}
 & Mafia  & “X looks good.” & \textit{hypo\_role\_in(sp,\,tg,\,bystander)} & $S(a)+=w_{mid}\,\mathbf 1[\text{bystander}(tg)]$\\
\cline{2-5}
 & Mafia  & YES vote to lynch & \textit{hypo\_role\_in(voter,\,target,\,mafia)} & $S(a)+=w_{low}\,\mathbf 1[\text{mafia}(target)]$\\
\hline
\end{tabular}
\caption{Constraint catalogue across datasets.  \emph{Constraint Format} shows the abstract grammar used by the extractor;  \emph{Mathematical Representation} shows how each constraint modifies the score $S(a)$ of an assignment $a$.  Assertions multiply $S$ by a large factor $w_A\!\gg\!1$, whereas hypotheses add small weights $w_{low}$, $w_{mid}$ and $w_{high}$(dataset-specific, all less than 1).}
\label{table:constraint_summary}
\end{table*}

\subsection{CSP-Based Role Inference Mechanism}

Given the generalized constraint framework introduced previously, we now describe the role inference process in detail. Our approach leverages Constraint Satisfaction Problems (CSPs) and a specialized scoring function to estimate posterior probabilities of player roles.

\subsubsection{Constraint Satisfaction Framework}

Formally, we define the CSP-based role inference as follows. An SDG-CSP model at time \(t\) is a tuple:
$
\mathcal{M}_t = (P, R, \mathcal{C}_t),
$
where \(P\) is the player set, \(R\) is the role set, and \(\mathcal{C}_t\) is the accumulated constraint set consisting of evidence (E), phenomenon (P), assertions (A), and hypotheses (H).

The feasible assignment space at time \(t\), denoted as \(AS_t\), consists of all assignments \(a: P \to R\) satisfying the accumulated hard constraints (\(E \cup P\)):
\[
AS_t = \{a \mid a \models E, a \models P \}.
\]

\subsubsection{Assignment Filtering via Hard Constraints}

Each new evidence or phenomenon constraint reduces the feasible assignment space. Formally, for any constraint \(c \in E \cup P\), the set of feasible assignments is updated by:
\[
AS_{t+1} = \{ a \in AS_t \mid a \models c \}, \quad \text{where } AS_{t+1} \subseteq AS_t.
\]

Thus, hard constraints monotonically reduce the feasible set, ensuring consistency and correctness in role inference.

\subsubsection{Scoring Function with Weighted Soft Constraints}

To incorporate player-driven constraints (Assertions and Hypotheses), we introduce a novel scoring function. Given an assignment \(a\), the scoring function is defined as:
\begin{equation}\label{eq:score_function}
S(a)=\left(\prod_{c \in A, a\models c} w_A\right) \cdot \left(1+\sum_{h \in H, a\models h} w_H(h)\right),
\end{equation}
where:
\begin{itemize}
    \item \(w_A \gg 1\) is the high weight assigned to assertions, strongly motivating their satisfaction.
    \item \(w_H(h)\) is the weight of hypotheses, calculated either manually or via Information Gain (IG): $w_H(h)=\text{IG}(h)$
\end{itemize}

To estimate posterior probabilities, we normalize these scores over the entire feasible set \cite{cover1999elements}:
\begin{equation}\label{eq:posterior_prob}
    Pr(a|\mathcal{C}_t)=\frac{S(a)}{\sum_{a' \in AS_t} S(a')}.
\end{equation}

\subsubsection{Information Gain Weighting}

We adopt information-theoretic principles to dynamically adjust hypothesis weights. For hypothesis \(h\), the information gain (IG) is calculated as the reduction in entropy when incorporating \(h\):
\begin{equation}\label{eq:ig_formula}
\text{IG}(h)=H(\text{prior}) - H(\text{posterior}|h),
\end{equation}
where entropy
$
H(X)=-\sum_{x}Pr(x)\log Pr(x).
$

The IG-based weighting measures how informative a hypothesis is, assigning greater weight to hypotheses that significantly clarify role distributions. When all assertions are truthful, using high–weight soft constraints is (almost) equivalent to treating them as hard constraints: the posterior probability error is bounded by ${\lvert Pr_s(a)-Pr_h(a)\rvert}\le 1/{w_A}$ (Formal proof in \textbf{Appendix B}).

\subsubsection{Inference Procedure and Computational Complexity}
Given the constraint set $\mathcal{C}_t$, we proceed in three steps:
\begin{enumerate}
    \item \textbf{Prune}  Apply all \emph{hard} constraints $(E\!\cup\!P)$ to eliminate infeasible worlds and obtain the candidate set $AS_t$.
    \item \textbf{Score \& normalize}  For every $a\in AS_t$ (or for a Monte-Carlo sample thereof), compute the soft–weighted score $S(a)$ from Eq.~\eqref{eq:score_function} and obtain the normalized posterior $Pr(a\mid\mathcal{C}_t)$ via Eq.~\eqref{eq:posterior_prob}.
    \item \textbf{Marginals and MAP}  Derive (i) player–wise marginal posteriors by summing $Pr(a\mid\mathcal{C}_t)$ over worlds that assign a given role, and (ii) the maximum–a-posteriori world $\hat a=\arg\max_{a\in AS_t} Pr(a\mid\mathcal{C}_t)$.
\end{enumerate}

For larger lobbies, we \emph{could} replace step~2 with an MCMC sampler \cite{andrieu2008tutorial}: starting from a feasible world, role‑swap proposals would respect global role counts and mixing could be sped up by caching partial soft‑scores.  With $M$ samples this strategy would scale as $\mathcal{O}(M n)$, give an unbiased estimator of all marginals, and take the highest‑scoring sample as MAP.  Because current SDG datasets are low‑dimensional ($n{\le}10$), we did not deploy this variant; testing it is left to future work on highly heterogeneous games such as \emph{Blood on the Clocktower}.

\section{Dataset}

\subsection{Game Overview}

\paragraph{Avalon}
Players are secretly assigned to \textit{good} roles (Merlin, Percival, Servants) or \textit{evil} roles (Morgana, Assassin).  Each round follows the fixed loop \emph{team proposal $\rightarrow$ vote $\rightarrow$ quest resolution}; after three successful quests the Assassin may attempt to identify Merlin for a last-chance victory.

\paragraph{Mafia}
Roles are \textit{Mafia} (evil) and \textit{Bystanders} (good).  
The game alternates \emph{night} (Mafia privately eliminate one target) and \emph{day} (public discussion + lynch vote) cycles until either faction is wiped out.

\subsection{Datasets Description}

We evaluate our approach on three datasets:

\begin{itemize}
    \item \textbf{Avalon NLU Dataset} \cite{stepputtis2023long}: 21 games (6 players each), detailed dialogues, votes and quests.
    \item \textbf{Mafia Dataset} \cite{ibraheem2022putting}: 44 games (4-10 players, 460 participants), detailed dialogues, votes, and night eliminations.
    \item \textbf{AvalonLogs Dataset} \cite{avalonlogs}: Large-scale dataset with 12,699 Avalon games (5-10 players), recording quests, votes, roles, and assassination choices.
\end{itemize}

Table~\ref{table:constraint_summary} lists the full catalogue with the corresponding mathematical semantics used by the solver. We manually annotate Dataset~1 and Dataset~2 as Truth or Lie, where Lie games have at least one explicitly false role claim from a good-aligned player (details in \textbf{Appendix C}).

\subsection{Dataset Preprocessing and Constraint Extraction via LLM}

For every game log we (i) \textbf{segment} the transcript into temporal blocks—quests for Avalon, day/night cycles for Mafia; (ii) \textbf{annotate} each block with LLMs (gpt-4.1-mini) using a few-shot prompt that maps every utterance or event to one of four first-order constraint schemas; the model outputs a JSON list of predicate instances; (iii) \textbf{post-process} the JSON by normalising player IDs, deduplicating constraints, enforcing type checks, and dropping ill-formed entries.  We then feed the resulting constraint sets sequence directly to the CSP solver. AvalonLogs is processed automatically with the same pipeline.  A rigorous, dual-author audit on Datasets 1 and 2 validated our constraint extraction, confirming 100\% fidelity by tracing all hard constraints to the raw logs and ensuring the inclusion of all mandatory events.
(See the complete prompt templates in \textbf{Appendix F}.)
\section{Experiments}

In this section we describe our experimental protocol and evaluation results across three datasets (Avalon–NLU, Mafia, AvalonLogs).  We compare three classes of methods:
\begin{itemize}
  \item \textbf{CSP-only:} purely constraint‐based inference under five settings (\emph{Strict}, \emph{+Assert}, \emph{+HypIG}, \emph{+HypM}, \emph{+TurnIG}).
  \item \textbf{LLM-only:} direct probability estimation by prompting a LLM on dialogue history (global vs.\ turn‐based).
  \item \textbf{LLM+CSP:} LLM predictions augmented with CSP posterior or MAP ``assist'' under the same five CSP settings.
\end{itemize}

We evaluate three {\em perspectives}: the {\em objective} view (public evidence only), each {\em good} role’s private view (Merlin, Percival, Servant, Bystander), and each {\em evil} role’s view (Assassin, Mafia). For AvalonLogs—which lacks dialogue—we run only the full-information variants (LLM-Global and LLM,+CSP,+HypIG) and sample one quest per game to keep the large corpus tractable. 

Unlike prior work that merges roles into broad classes, we report exact per-player role accuracy, a much harder target. Potential baselines like DeepRole~\cite{serrino2019finding} and SVC-Assassin~\cite{chuchro2022training} are not included: DeepRole optimizes win-rate through full-game policy learning, SVC-Assassin predicts only the final assassination, and both models ignore dialogue, hard-code the variants, and depend on rule-specific network engineering, making them incomparable to our round-level, multi-dataset role-inference protocol. For the same reasons, we also forgo dataset‑specific fine‑tuning of the LLM backbones: the corpora are too small for reliable supervised adaptation, instruction‑tuned models are the standard baselines in previous works, and preserving out‑of‑domain generalization is central to our evaluation.

\begin{table*}[t]
\centering
\small
\setlength{\tabcolsep}{4pt}
\begin{tabular}{|l|l|c|*{6}{c}|*{3}{c}|}
\hline
\multirow{2}{*}{\textbf{Family}} &
\multirow{2}{*}{\textbf{Setting}} &
\multirow{2}{*}{\textbf{Cond.}} &
\multicolumn{6}{c|}{\textbf{Avalon‑NLU}} &
\multicolumn{3}{c|}{\textbf{Mafia‑UCB}} \\
\cline{4-12}
 & & & Obj & MRL & PCV & ASN & MRG & LSV & Obj & BYS & MAF \\
\hline
\multirow{5}{*}{CSP}
 & Strict   & T & 0.28/0.24 & 0.63/0.60 & 0.69/0.67 & 0.59/0.53 & 0.59/0.53 & 0.56/0.53 & 0.70/0.90 & \textbf{0.72/0.91} & \textbf{1.00/1.00} \\
 & +Assert  & T & 0.33/0.29 & \textbf{0.65/0.60} & 0.75/0.73 & \textbf{0.61/0.53} & \textbf{0.61/0.53} & 0.59/0.60 & 0.70/0.90 & 0.72/0.91 & 1.00/1.00 \\
 & +HypIG   & T & \textbf{0.34/0.30} & 0.65/0.60 & 0.76/0.79 & 0.61/0.53 & 0.61/0.53 & \textbf{0.60/0.65} & \textbf{0.73/0.76} & \textbf{0.74/0.76} & 1.00/1.00 \\
 & +HypM    & T & \textbf{0.33/0.31} & 0.65/0.60 & 0.75/0.79 & 0.61/0.53 & 0.61/0.53 & 0.60/0.65 & 0.73/0.76 & 0.74/0.76 & 1.00/1.00 \\
 & +TurnIG  & T & 0.34/0.29 & 0.65/0.60 & \textbf{0.76/0.80} & 0.61/0.53 & 0.61/0.53 & 0.60/0.63 & 0.72/0.79 & 0.73/0.79 & 1.00/1.00 \\
\hline
\multirow{2}{*}{LLM}
 & GChat & T & 0.14/0.20 & 0.20/0.24 & 0.16/0.19 & 0.13/0.19 & 0.16/0.25 & 0.13/0.21 & 0.56/0.62 & 0.56/0.61 & 0.59/0.77 \\
 & TChat & T & 0.11/0.12 & 0.21/0.25 & 0.14/0.18 & 0.11/0.12 & 0.16/0.20 & 0.11/0.16 & 0.66/0.72 & 0.61/0.66 & 0.67/0.82 \\
\hline
\multirow{5}{*}{\shortstack{LLM\\+\\CSP}}
 & Strict   & T & 0.16/0.12 & 0.41/0.39 & 0.44/0.41 & 0.42/0.39 & 0.42/0.39 & 0.12/0.17 & 0.69/0.89 & 0.70/0.89 & 0.89/0.93 \\
 & +Assert  & T & 0.19/0.15 & 0.42/0.40 & 0.47/0.45 & 0.44/0.40 & 0.44/0.40 & 0.13/0.18 & 0.69/0.89 & 0.70/0.89 & 0.89/0.93 \\
 & +HypIG   & T & 0.19/0.16 & 0.42/0.40 & 0.48/0.50 & 0.44/0.40 & 0.44/0.40 & 0.12/0.19 & 0.69/0.74 & 0.70/0.74 & 0.89/0.93 \\
 & +HypM    & T & 0.19/0.16 & 0.42/0.40 & 0.47/0.50 & 0.44/0.40 & 0.44/0.40 & 0.13/0.20 & 0.69/0.74 & 0.70/0.73 & 0.89/0.93 \\
 & +TurnIG  & T & 0.19/0.15 & 0.42/0.40 & 0.47/0.45 & 0.44/0.40 & 0.44/0.40 & 0.11/0.15 & 0.69/0.86 & 0.70/0.86 & 0.92/0.95 \\
\hline
\multirow{1}{*}{Random}
 &  --  & -- & 0.2667 & 0.4 & 0.4 & 0.5833 & 0.5833 & 0.3333 & 0.68 & 0.6889 & 1.00 \\
\hline
\end{tabular}

\caption{Truthful‑Good Results on dialogue datasets (\textbf{Lying-Good results in Appendix G}).  
Each cell reports \emph{marginal accuracy} / \emph{MAP accuracy}.  
\textbf{Strict}: evidence \& phenomenon only; \textbf{+Assert}: adds assertions;  
\textbf{+HypIG}: global hypotheses (IG weights); \textbf{+HypM}: global hypotheses (manual);  
\textbf{+TurnIG}: turn-local hypotheses;  
\textbf{GChat}: LLM with global chat + state; \textbf{TChat}: LLM with turn chat, state.  
Role abbreviations—Avalon: MRL Merlin, PCV Percival, ASN Assassin, MRG Morgana, LSV Loyal Servant;  
Mafia: BYS Bystander, MAF Mafia; Obj = objective.
Only the first highest value per metric is highlighted in bold.}
\label{tab:dialogue_results}
\end{table*}

\subsection{Methods and Settings}
\paragraph{CSP‐only } We accumulate \emph{evidence} and \emph{phenomenon} constraints and evaluate under five settings:
\begin{enumerate}
  \item \emph{Strict}: evidence + phenomenon only.
  \item \emph{+Assert}: adds high‐weight assertions (carried forward).
  \item \emph{+HypIG}: adds all hypotheses with IG weights.
  \item \emph{+HypM}: adds all hypotheses with manual weights. 
  
  (grid‑search tuned in experiments; details in Appendix~I)
  \item \emph{+TurnIG}: adds current‐turn hypotheses with IG weights.
\end{enumerate}

\paragraph{LLM‐only} We prompt the LLM with the chat history up to the current round (either global or turn‐only) and the public event log of quests or eliminations, asking it to return for every player a probability distribution over roles together with a MAP assignment that respects the known role counts.

\paragraph{LLM+CSP} Identical to LLM‐only but the prompt is augmented with the CSP posterior table and MAP assignment under one CSP setting (we experiment with all five).

\subsection{Views}
Let $V$ be the set of all perspectives:
$
V = \{\text{objective}\}\;\cup\;\{\text{each role}\}.
$
For each $v\in V$ we add ``perspective evidence'':
If $v=\text{Merlin}$, add hard constraints that all evils are known;
if $v=\text{Percival}$, constrain the Merlin/Morgana candidates to those roles;
if $v$ is evil, reveal fellow evils;
otherwise, none.

\subsection{Evaluation Metrics} 

For each round $q$ of a game we evaluate two quantities:

\paragraph{Marginal Accuracy}  
Let $P$ be the number of players, $r_p^{\star}$ the ground-truth role of player $p$, and $a_{p,r}$ the model’s posterior for role $r$ is:  
$
\mathrm{MA}_q \;=\; \frac1P \sum_{p=1}^{P} a_{p,r_p^{\star}} .
$

\paragraph{MAP Accuracy}  
Let $\hat r_p$ be the role assigned to $p$ in the model’s maximum–a-posteriori (MAP) world, then:
$
\mathrm{MAP}_q \;=\; \frac1P \sum_{p=1}^{P} \mathbf 1\!\bigl[\hat r_p = r_p^{\star}\bigr].
$

\paragraph{Aggregation}  
Both metrics are first averaged over the $Q$ rounds of a single game, then across the set $\mathcal G$ of games in a dataset
$
\text{Overall}\;m \;=\; \frac{1}{|\mathcal G|} \sum_{g\in\mathcal G} \; \frac1Q \sum_{q=1}^{Q} m_q $ where $
m\in\{\mathrm{MA},\mathrm{MAP}\},
$
reporting the mean and standard deviation.

\subsection{Experimental Protocol}
For Avalon–NLU and Mafia we run all methods × settings × views at every round.  For AvalonLogs we sample a single quest per game and run CSP-only (5 settings), LLM‐only (global) and LLM+CSP (\emph{HypIG}). Scaling all weights by factor $\lambda\!\in\!\{0.25,0.5,1,2,4\}$ set altered MAP accuracy by $\leq$ 1.5 pp (percentage points), so we retain the default ($w_{strong}$, $w_{mid}$, $w_{low}$) = (0.5,0.2,0.1) weights (tuned in \textbf{Appendix~I}) for all reported results.  In addition, on Avalon–NLU we conduct supplementary \emph{LLM‐only} experiments to compare pure dialogue‐based inference across multiple LLMs (GPT-4o-mini, GPT-4o, GPT-4.1, Deepseek-v3, Gemini-2.0-flash). Details are presented in \textbf{Appendix D}.

\section{Results}

\begin{figure*}[!t]
\centering
\subfloat[Marginal accuracy by quest (servant view)]{%
\includegraphics[width=0.5\textwidth]{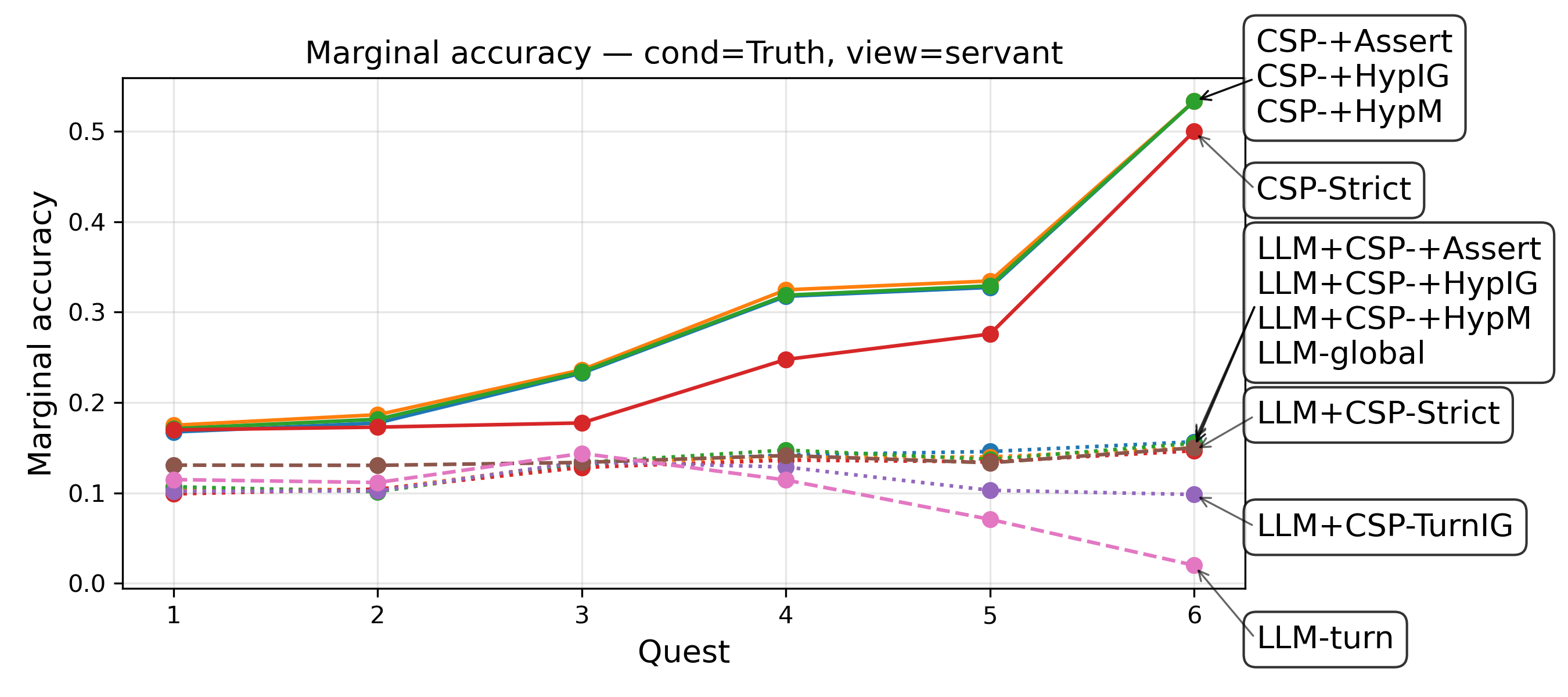}%
\label{fig:marginal-trend}%
}\hfill
\subfloat[MAP accuracy by quest (servant view)]{%
\includegraphics[width=0.5\textwidth]{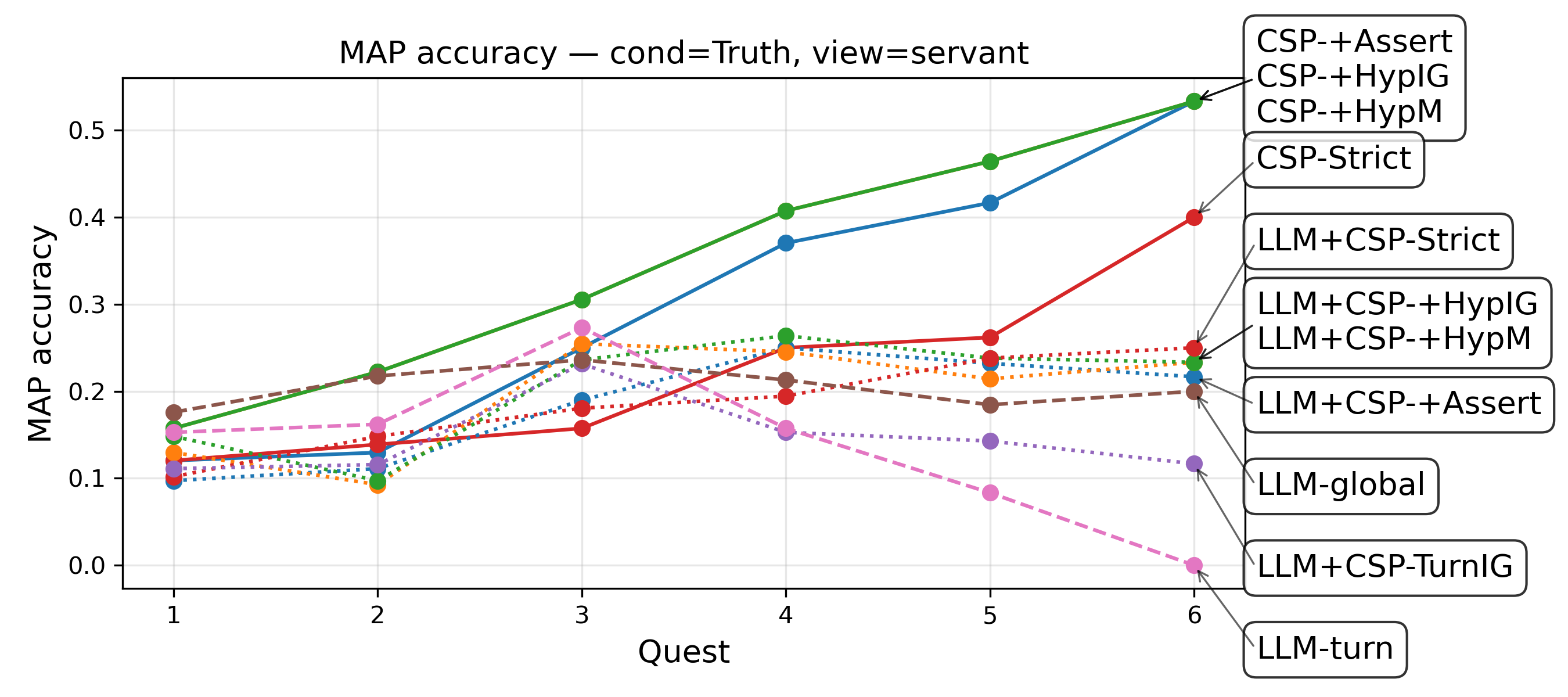}%
\label{fig:map-trend}%
}
\caption{Quest-by-quest accuracy trends on Avalon-NLU. Quest 6 = assassination round following three successful quests.}
\label{fig:quest-trends}
\end{figure*}

\begin{table*}[t]
\centering
\small
\begin{tabular}{|l|l|*{6}{c}|*{6}{c}|}
\hline
\multirow{2}{*}{\textbf{Family}} &
\multirow{2}{*}{\textbf{Setting}} &
\multicolumn{6}{c|}{\textbf{AvalonLogs — 6‑player}} &
\multicolumn{6}{c|}{\textbf{AvalonLogs — 9‑player}} \\
\cline{3-14}
 & & Obj & MRL & PCV & MRG & LSV & EVM & Obj & MRL & PCV & MRG & LSV & EVM \\
\hline
\multirow{2}{*}{CSP}
 & Strict   & 0.290 & 0.614 & 0.635 & 0.584 & 0.390 & 0.691 & 0.302 & 0.595 & 0.568 & 0.554 & 0.348 & 0.614 \\
 & +HypIG   & 0.326 & 0.615 & 0.637 & 0.577 & 0.419 & 0.729 & 0.360 & 0.666 & 0.525 & 0.578 & 0.403 & 0.674 \\
\hline
\multirow{1}{*}{LLM}
 & Global   & 0.283 & 0.360 & 0.325 & 0.298 & 0.290 & 0.464 & 0.280 & 0.360 & 0.309 & 0.313 & 0.291 & 0.417 \\
\hline
\multirow{1}{*}{LLM+CSP}
 & +HypIG   & 0.289 & 0.613 & 0.633 & 0.584 & 0.389 & 0.691 & 0.301 & 0.593 & 0.564 & 0.554 & 0.347 & 0.614 \\
\hline
\end{tabular}

\vspace{-4pt}
\caption{AvalonLogs performance split by game size.  
Each cell shows \emph{marginal accuracy}.  
Roles: Obj (objective), MRL (Merlin), PCV (Percival), MRG (Morgana),  
LSV (loyal–servant aggregate), EVM (evil–minion aggregate).  
The 6-player subset illustrates a small-scale scenario, while the 9-player subset
represents a larger game.  In both cases CSP-based methods outperform a pure
LLM baseline, and the hybrid \emph{LLM+CSP} variant retains most CSP gains
while leveraging LLM reasoning.}
\label{tab:avalonlogs_results}
\end{table*}

\subsection{Dialogue Datasets Analysis}

Table~\ref{tab:dialogue_results} summarizes the results for both the Avalon-NLU and Mafia-UCB datasets under various conditions. Because the Lying‑Good split contains only a handful of games (n = 4), we regard its findings as preliminary; the complete L‑condition table is provided in \textbf{Appendix G}. Based on both sets of results, several clear observations emerge:

\paragraph{CSP methods dominate consistently} Regardless of whether good players lie, CSP methods achieve the highest overall accuracy. Specifically, CSP+HypIG and CSP+HypM settings yield the best performance, with HypIG marginally outperforming HypM due to information-gain weighting. Interestingly, the CSP+TurnIG setting, which incorporates turn-local hypotheses, surpasses CSP+Assert, and CSP+Assert is superior to CSP-Strict. In certain viewpoints, however, CSP+Assert already attains the column‑wise optimum, and adding further soft cues does not raise—but also never lowers—the score, making the extra layers a pure bonus.
 This trend indicates the incremental benefit of adding soft information in structured formats. All reported improvements of CSP over the LLM baseline are statistically significant according to two‐sided Wilcoxon signed–rank and paired $t$‑tests ($p<0.05$); see \textbf{Appendix E} for test details.

\paragraph{LLM struggles with complex reasoning} Pure LLM approaches significantly underperform compared to CSP in the more complex Avalon-NLU dataset, indicating difficulties in precise role deduction from dialogue alone. Introducing CSP-derived information (LLM+CSP methods) slightly improves overall accuracy, especially from special-role viewpoints (Merlin, Percival, Morgana, Assassin). However, from the objective viewpoint and Loyal Servant perspective, accuracy gains are minimal or even negative, reflecting that limited structured knowledge may mislead the LLM.

\paragraph{Simpler dialogue benefits LLM} For the simpler Mafia-UCB dataset, performance differences between LLM and CSP methods narrow significantly. Here, the TChat setting (turn-based contexts) notably outperforms the global-context setting (GChat). Moreover, under the CSP-supported HypIG condition, the LLM+CSP approach even surpasses pure CSP, illustrating the synergy possible when combining structured priors and simpler dialogue contexts.

However, a critical limitation of the LLM becomes evident: despite having perfect information from the Mafia perspective, the pure LLM still fails to achieve full accuracy due to inherent hallucinations and contextual confusion. Even with 100\% correct CSP priors, the LLM struggles to disregard misleading or irrelevant dialogue elements, underscoring the necessity for explicit fine-tuning or stronger constraints to directly guide accurate reasoning.

\paragraph{Marginal vs. MAP accuracy} Interestingly, pure CSP consistently shows slightly higher marginal accuracy compared to MAP accuracy. The reverse pattern occurs for LLM methods, suggesting that while LLM confidently converges towards a single best solution, the CSP generates a more probabilistic distribution of roles, spreading uncertainty across multiple assignments.

\paragraph{Impact of Lying-Good behavior} In both Avalon‑NLU and Mafia‑UCB, only a few games contain explicit role‑lying from good players (\textbf{Appendix G}). In these limited cases we see anecdotal shifts: CSP accuracy sometimes moves in favor of good‑aligned perspectives, whereas our prompt‑based LLM baselines tend to drop slightly across viewpoints. Because the sample is too small for firm statistics, we treat these observations as illustrative only. A larger, dedicated study of deception effects is left for future work.

\begin{figure*}[!t]
\centering
\subfloat[Overall MAP accuracy by view type]{%
\includegraphics[width=.517\linewidth]{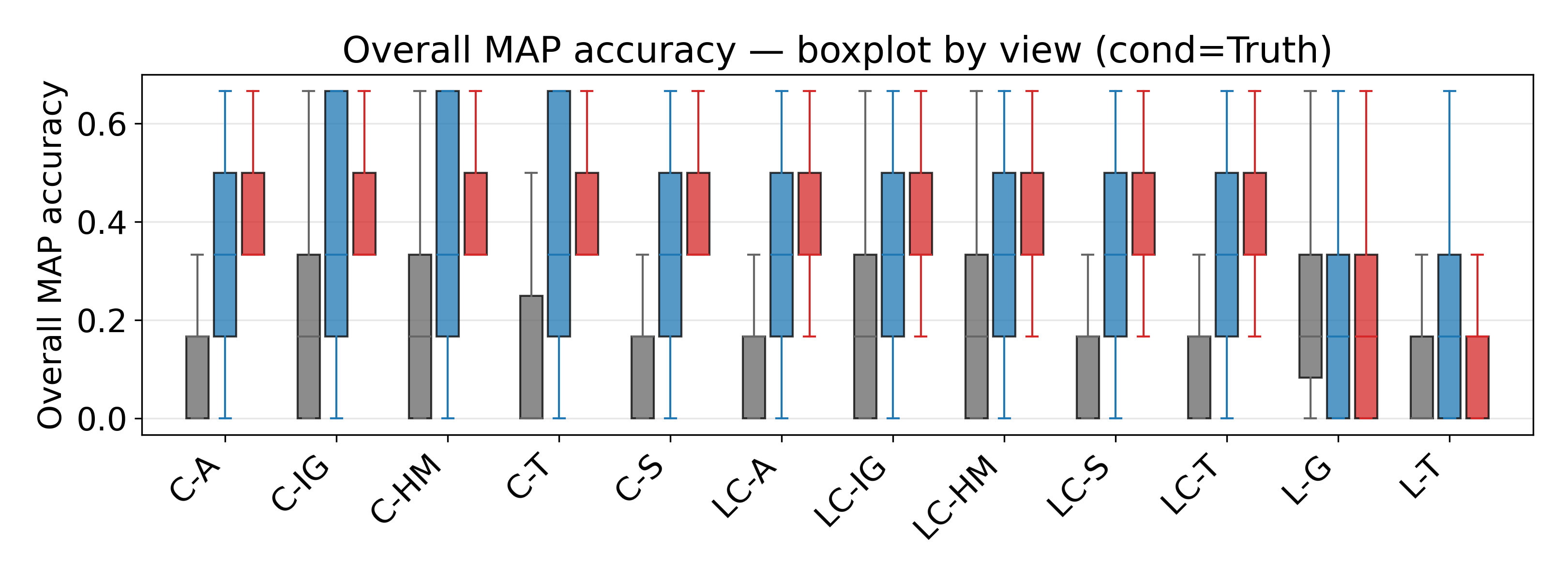}%
\label{fig:map-box}%
}\hfill
\subfloat[Overall marginal accuracy scatter plot by view type]{%
\includegraphics[width=.483\linewidth]{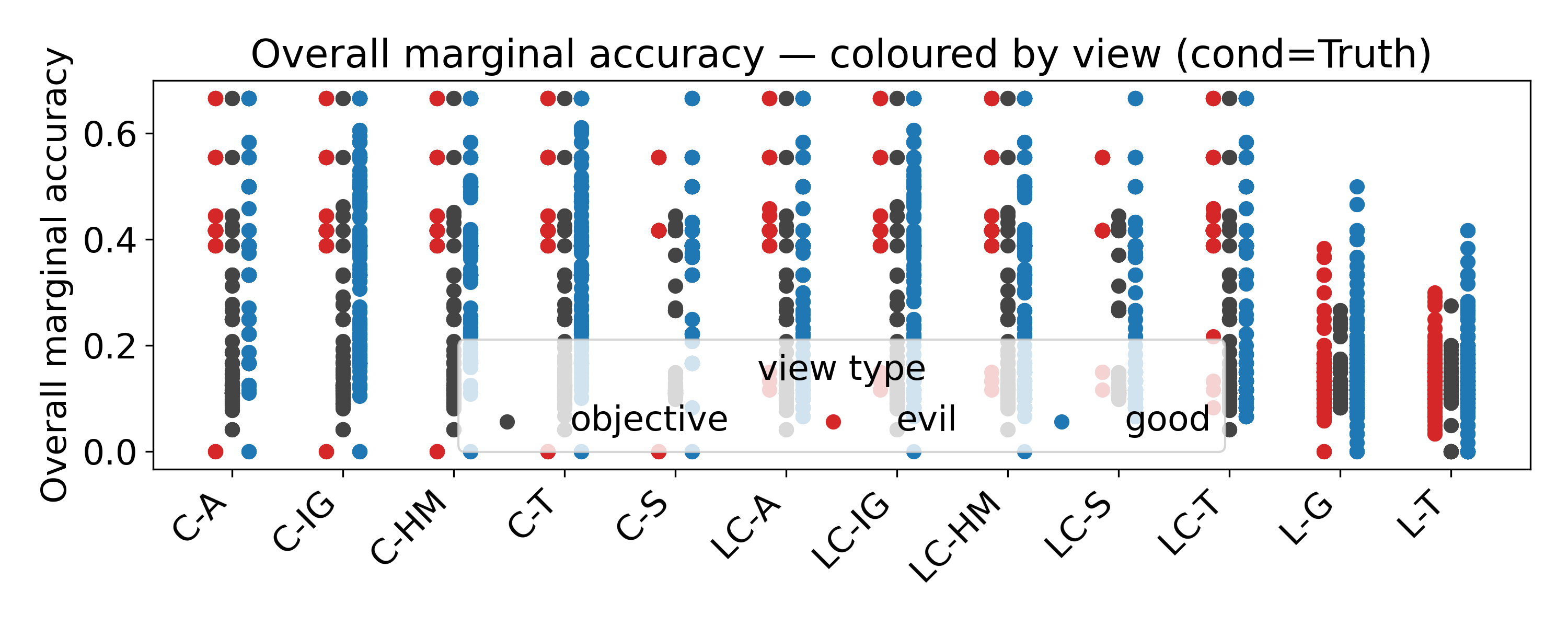}%
\label{fig:marginal-scatter}%
}
\caption{Accuracy distribution across different perspectives (\emph{Truthful} condition). Good-role perspectives (blue) achieve higher accuracy, evil-role perspectives (red) exhibit broader variance, and objective views (gray) record lower accuracy levels. Abbreviations:  
\textit{C} = CSP,\; \textit{LC} = LLM+CSP,\; \textit{L} = LLM;  
\textit{A} = \,+Assert,\; \textit{IG} = +HypIG,\; \textit{HM} = +HypM,\;  
\textit{T} = TurnIG,\; \textit{S} = Strict.}
\label{fig:box-scatter}
\end{figure*}

\begin{table*}[!t]
\centering\small
\setlength{\tabcolsep}{4.5pt}
\begin{tabular}{|l|ccccc|ccccc|}
\hline
\multirow{2}{*}{\textbf{Model}} 
  & \multicolumn{5}{c|}{\textbf{LLM-Global}} 
  & \multicolumn{5}{c|}{\textbf{LLM+CSP+HypIG}} \\
\cline{2-11}
 & Obj & ASN & PCV & MRG & LSV 
 & Obj & ASN & PCV & MRG & LSV \\
\hline
GPT-4o-mini
  & 0.14/0.20 & 0.13/0.19 & 0.16/0.19 & 0.16/0.25 & 0.13/0.21 
  & 0.19/0.16 & 0.44/0.40 & 0.48/0.50 & 0.44/0.40 & 0.12/0.19 \\
GPT-4o
  & 0.15/0.16 & 0.12/0.08 & 0.20/0.27 & 0.17/0.18 & 0.15/0.16 
  & 0.19/0.17 & 0.44/0.40 & 0.47/0.50 & 0.44/0.41 & 0.15/0.15 \\
GPT-4.1
  & 0.15/0.12 & 0.11/0.08 & 0.25/0.30 & 0.16/0.14 & 0.15/0.13 
  & 0.19/0.16 & 0.43/0.40 & 0.48/0.50 & 0.42/0.40 & 0.15/0.15 \\
DeepSeek-v3
  & 0.16/0.16 & 0.20/0.16 & 0.22/0.25 & 0.19/0.21 & 0.17/0.16 
  & 0.19/0.16 & 0.46/0.40 & 0.51/0.55 & 0.46/0.40 & 0.16/0.14 \\
Gemini-2.0-flash
  & 0.16/0.18 & 0.13/0.12 & 0.18/0.22 & 0.18/0.16 & 0.16/0.17 
  & 0.19/0.16 & 0.41/0.39 & 0.48/0.50 & 0.38/0.36 & 0.16/0.17 \\
\hline
\end{tabular}
\caption{Marginal/MAP accuracy on Avalon-NLU under pure LLM and LLM+CSP+HypIG. Merlin omitted (identical across setups; likely as Merlin's deterministic constraints heavily prune the domain, leading to uniform inferences across all models).}
\label{tab:llm_ablation_avalon_full}
\end{table*}

\subsection{No-dialogue Large Scale Dataset: AvalonLogs}

Without dialogue, AvalonLogs provides only quest and vote records. We give six‑ and nine‑player results here in Table~3 and place full tables in \textbf{Appendix H}. Only Strict and +HypIG settings are shown, because all other variants alter accuracy by $\leq 2$ pp, and +HypIG is typically the best. Under +HypIG setting, the objective score rises to 0.33 $\sim$ 0.36 , while role views climb further (Merlin 0.67, Evil‑Minion 0.73). These gains result mainly from event‑level assertions, augmented by IG‑weighted hypotheses.

LLM alone, confined to frequency heuristics, stays near 0.28 objectively and below 0.37 for special roles; CSP surpasses it by 5–8 pp on low‑information roles and 25–30 pp on informative ones. Feeding CSP back to the model yields the same scores as CSP-Strict, confirming that LLM performs better in no-dialogue dataset yet still relies on CSP for combinatorial reasoning. Compared with the dialogue‑rich Avalon‑NLU corpus, gains here are smaller because the evidence ceiling is lower, but CSP remains dominant. Accuracy also exhibits a size‑related pattern—dipping at 7 players, recovering at 8–9, and falling again at 10 (discussion in \textbf{Appendix~H}). Conversely, LLM and LLM+CSP degrade on Avalon‑NLU, showing that unfiltered dialogue can swamp the model even with accurate CSP hints. Overall,(1) mission and vote data alone give a solid but bounded signal, whereas dialogue unlocks the deeper role logic that makes SDGs dialogue‑driven; (2) LLMs excel at information extraction but not at the structured reasoning as CSP does.  

\subsection{Analysis of Trends and Viewpoints}

Figures~\ref{fig:marginal-trend} and \ref{fig:map-trend} show how accuracy evolves throughout the quests from the servant's viewpoint. CSP-based methods (solid lines) exhibit consistent improvement in accuracy, particularly from Quest 3 onward, sharply rising in the final assassination round (Quest 6). CSP variants utilizing soft constraints (+Assert, +HypIG, +HypM) lead performance, whereas the Strict setting lags slightly due to reliance solely on hard constraints. Pure LLM methods (dashed and dotted lines) have minimal improvements or even degrade over time, especially in the LLM-turn mode. Augmenting LLM with CSP posterior information (LLM+CSP) provides moderate gains but still falls short compared to pure CSP, highlighting the advantage of structured probabilistic reasoning.

Figure~\ref{fig:map-box} confirms that accuracy varies significantly by perspective, with evil-role views consistently achieving higher median accuracies across methods. Good-role perspectives, however, exhibit notable variance, reflecting their susceptibility to misleading or ambiguous dialogue. The objective perspective generally records the lowest accuracies, emphasizing that informed viewpoints substantially aid role prediction. Figure~\ref{fig:marginal-scatter} further highlights this trend, clearly separating good/evil-role predictions toward higher accuracy, while objective predictions scatter more broadly, indicating the nuanced challenges inherent to this perspective.

\subsection{Ablation: Different LLM Capabilities}

Across LLM backbones, from lightweight to state-of-the-art, increasing model scale yields only modest and sometimes negative returns on role-inference accuracy, as shown in Table 4. This suggests the performance bottleneck is not a matter of model capacity but a conceptual limitation, as larger models tend to amplify superficial heuristics rather than master the combinatorial reasoning essential for SDGs. Conversely, augmenting any LLM with CSP posteriors produces a uniform and significant accuracy uplift, particularly for roles with hidden information like Assassin, Percival, and Morgana. This consistent improvement demonstrates that the hybrid design successfully addresses a core reasoning deficit inherent in current language models. Therefore, the results indicate that structured, constraint-based reasoning is not merely a temporary fix for present models but will likely remain an essential component for achieving high-fidelity inference, underscoring the framework's long-term relevance even as foundation models continue to evolve.

\section{Conclusion}

In this work, we introduced CSP4SDG, a generalized probabilistic constraint-satisfaction framework for role inference in SDGs. We formulated role identification as a CSP by decomposing game information into structured, linguistically-agnostic constraints, and leveraged information-theoretic principles to dynamically weight soft constraints. Extensive evaluations across three public datasets with multiple CSP configurations, varying role perspectives, and LLM ablation experiments confirmed that pure LLMs are insufficient for human-level role identification tasks, and CSP4SDG consistently achieves superior accuracy and interpretability compared to pure LLM baselines and hybrid approaches. Future work will explore interactive, human-in-loop constraint processing, test on large and complex datasets, and extend CSP-driven posteriors from passive inference to active, real-time decision support within live gameplay.

\clearpage
\bibliography{aaai2026}
\nocite{*}

\appendix
\section{Appendix A: Related works}

\subsection{Role Identification in Social Deduction Games}

Social deduction games such as Mafia and Avalon are formalized as partial-information games where players infer hidden roles through public interactions \cite{braverman2008mafia}. Early work extracted lexical and pragmatic cues of deception from an online Chinese Mafia corpus \cite{zhou2008cues}, while subtle shifts in politeness and reciprocity were shown to presage betrayal in Diplomacy email dialogues \cite{niculae2015linguistic}. The release of large-scale corpora like Mafiascum enabled supervised role-classification models \cite{de2018mafiascum}. More recently, fine-tuned transformer models have been applied to Mafia chat logs \cite{ibraheem2022putting}, and LLMs were benchmarked on long-horizon Avalon dialogues \cite{stepputtis2023long}. Multimodal datasets further demonstrated the benefit of combining textual and visual cues for persuasion strategy in Werewolf \cite{lai2023werewolf}. Complementary non-linguistic approaches include DeepRole’s reinforcement-learning inference from voting records in Avalon \cite{serrino2019finding}, psychologically-inspired Werewolf agents using multi-perspective mental-state models \cite{nakamura2016constructing}, and an SVM-based Assassin classifier for Avalon’s endgame \cite{chuchro2022training}. Hybrid strategic–language frameworks such as Cicero for Diplomacy integrate planning algorithms with LLM-driven negotiation to achieve human-level performance \cite{meta2022human}. Other works explore SDG research in multiple dimensions \cite{kopparapu2022hidden, eger2018keeping, chi2024amongagents, sarkar2025training, kim2024fine, velikov2021rlerewolf, martinenghi2024llms, wu2024enhance, carminati2023hidden}. Despite these advances, existing methods often either neglect free-form dialogue, lack logical consistency in their inferences, or do not incorporate active information-theoretic decision criteria. In contrast, our CSP4SDG framework requires no model training or annotated datasets, relying solely on objective probabilistic reasoning and information metrics: it uses LLM as an information converter to extract structured ``if–then'' constraints, a CSP solver to enforce a coherent belief space, and entropy-based measures to guide prediction and action selection in real time.

\subsection{Constraint Programming \& Information-Theoretic Reasoning}

Constraint satisfaction views a problem as variables, finite domains, and hard relations that must all hold.  Decades of work on propagation and back-tracking make classical CSPs an exact and interpretable tool for combinatorial search \cite{freuder2006constraint}.  Soft extensions attach costs to violations (valued/weighted CSPs) so that infeasible but nearly consistent assignments can still be ranked \cite{schiex1995valued}.  Probabilistic (mixed) CSPs push this further by treating some variables as stochastic and maximising the likelihood that constraints are satisfied \cite{fargier1996mixed}.  These variants remain powerful for puzzles and scheduling, yet they presume that (i) constraints are fully specified \emph{a priori}, and (ii) all soft weights are hand-tuned—assumptions violated in social–deduction games, where constraints must be extracted from noisy language and weighted by their evidential reliability.

Information theory quantifies that reliability.  Entropy measures belief uncertainty; its expected reduction, or \emph{information gain} (IG), underlies decision-tree splits and active-learning heuristics \cite{mackay1992information}, and is framed as the \emph{value of information} (VOI) for rational Bayesian decision making.  We merge these ideas: conversational evidence becomes soft CSP constraints whose weights are proportional to their VOI, so highly informative utterances dominate the score while noisy chatter is automatically down-weighted.  This VOI-weighted probabilistic CSP yields exact per-turn role probabilities, preserves interpretability, and supplies structured inputs that further boost large language models—bridging symbolic, probabilistic and neural reasoning in social-deduction domains. 

Our IG-weighted formulation is conceptually close to valued or VOI CSP~\cite{schiex1995valued,mackay1992information}, yet those classical models rely on a fixed, hand-tuned cost matrix that must be re-engineered for every game variant.  To avoid such domain-specific tuning, we treat them as theoretical precedents rather than empirical baselines.

\section{Appendix B: Proof of Equivalence between Soft and Hard Assertions}

\begin{theorem}
Suppose all assertion constraints in $A$ are truthful, and let 
\[
S_s(a)=w_A^{|A|}\bigl(1+\sum_{h\in H,a\models h}w_H(h)\bigr)
\]
be the soft‐constraint score, with $w_A>1$, and 
\[
S_h(a)=1+\sum_{h\in H,a\models h}w_H(h)
\]
the score when assertions in $A$ are enforced as hard constraints.  Define posterior distributions
\[
Pr_s(a)=\frac{S_s(a)}{\sum_{a'\in A_t}S_s(a')},\quad
Pr_h(a)=\frac{S_h(a)}{\sum_{a'\in A_{all}}S_h(a')},
\]
where $A_{all}\subseteq A_t$ is the subset of assignments satisfying \emph{all} assertions.  Then for every assignment $a$,
\[
\bigl\lvert Pr_s(a)-Pr_h(a)\bigr\rvert \;\le\; \frac{1}{w_A}\,. 
\]
\end{theorem}

\begin{proof}
Partition the denominator of $Pr_s$ into contributions from 
$A_{all}$ and its complement.  For any $a\in A_{all}$,
\[
S_s(a)=w_A^{|A|}S_h(a),
\]
while for any $a'\notin A_{all}$, since it violates at least one assertion,
\[
S_s(a')\;\le\; w_A^{|A|-1}\bigl(1+\sum_{h\in H}w_H(h)\bigr)
\;\le\; \frac{w_A^{|A|}}{w_A}\,M,
\]
where $M=1+\sum_{h\in H}w_H(h)$.  Hence
\[
\sum_{a'\notin A_{all}}S_s(a') 
\;\le\; \frac{1}{w_A}\,w_A^{|A|}\,M 
\;\ll\;\sum_{a''\in A_{all}}w_A^{|A|}S_h(a'')
\]
for large $w_A$ (e.g. 100).
Thus
\[
Pr_s(a)
=\frac{w_A^{|A|}S_h(a)}{w_A^{|A|}\sum_{a''\in A_{all}}S_h(a'')
      \;+\;\sum_{a'\notin A_{all}}S_s(a')}
      \]\[
\;\approx\;
\frac{S_h(a)}{\sum_{a''\in A_{all}}S_h(a'')}=Pr_h(a)\,.
\]
A straightforward bounding of the residual term shows 
$\lvert Pr_s(a)-Pr_h(a)\rvert\le1/w_A$, completing the proof.
\end{proof}

\section{Appendix C: Dataset}

\subsection{Game Overview}

\paragraph{Avalon:}
Avalon is a social deduction game with hidden roles, deception, and cooperation. Players are divided into two factions: the loyal servants of Arthur (Good) and the minions of Mordred (Evil). Each round: (i) a leader proposes a team; (ii) the table votes to approve or reject; (iii) the approved team secretly submits Success/Fail cards; the quest succeeds if enough Success cards are played. Good wins with three successful quests; Evil wins with three failed quests or by assassinating Merlin after the fifth quest. Common special roles used across playgroups include Merlin (knows Evil) and Assassin (may attempt the final assassination); other roles such as Percival and Morgana are optional in many setups.%
\footnote{Official rulebook by the publisher Indie Boards \& Cards: \url{https://cdn.1j1ju.com/medias/a6/dc/c1-the-resistance-avalon-rulebook.pdf}; publisher game page: \url{https://indieboardsandcards.com/our-games/the-resistance-avalon/}.}
In our experiments, the public information we log includes team proposals, table votes, and quest outcomes; free-form dialogue provides claims and hypotheses that we convert into soft constraints.

\paragraph{Mafia:}
Mafia alternates Night and Day. At Night, the informed minority (Mafia) secretly selects a target. At Day, players discuss and then vote to eliminate a suspect. Town (uninformed majority; Bystanders in our cases) wins if all Mafia are eliminated; Mafia wins upon parity. Rulesets vary across communities; the canonical baseline is Dmitry (Dimma) Davidoff's original rules, with modern Werewolf variants extending role lists and procedures.%
\footnote{Original ``Mafia'' rules by the game's creator, Dmitry (Dimma) Davidoff: \url{https://gusandco.net/wp-content/uploads/2021/04/The-Original-Mafia-Rules-1.pdf}.}
For our datasets, we rely on publicly observable events (day votes, eliminations, night outcomes) as hard constraints, and we extract soft constraints from player dialogue claims.

\subsection{Datasets Description}

\paragraph{Dataset 1 (Avalon NLU Dataset):} This dataset \cite{stepputtis2023long} comprises 21 games with 6-player configurations, featuring Merlin, Percival, Morgana, Assassin, and Servants. The dataset records comprehensive dialogue logs, voting behaviors, team proposals, quest outcomes, and final role assignments. The dialogues extensively reflect long-term strategic reasoning and interactions.

\paragraph{Dataset 2 (Mafia Dataset):} Collected from a controlled environment via Amazon Mechanical Turk \cite{ibraheem2022putting}, this dataset includes 44 Mafia games with 4-10 players per game (approximately 460 participants total). Roles include Mafia and Bystanders, with detailed logs capturing daytime dialogues, nighttime Mafia communications, voting behaviors, and eliminations. Dialogues and interactions reveal deception strategies used by Mafia players.

\paragraph{Dataset 3 (AvalonLogs Dataset):} Sourced from AvalonGame.Online, this extensive dataset \cite{avalonlogs} contains 12699 Avalon games with varying player counts (5-10 players). Board configurations depend on player count. Data includes detailed game logs, team proposals, voting patterns, quest outcomes, role assignments, and assassination choices.

\paragraph{Truth/Lie annotation:}
For \textbf{Dataset~1} (Avalon-NLU) and \textbf{Dataset~2} (Mafia) we manually distinguish two evaluation splits. A game belongs to the \texttt{Lie} split iff \emph{at least one good-aligned player explicitly utters a false hidden-role claim} (\textit{e.g.}, Merlin claiming``I am Percival.''). Otherwise it is placed in \texttt{Truth}. In Avalon such lies (2 samples) predominantly arise from Merlin who pretend to be Percival; in Mafia (2 samples) every Bystander has no special power, hence there is no strategic incentive to lie, yet a small fraction of Bystanders still claim to be Mafia for fun—we likewise regard these as lies.

\subsection{Dataset Preprocessing and Constraint Extraction via LLM}

Our preprocessing pipeline consists of three stages: \textbf{(i)~Segmentation} of raw logs into coherent temporal blocks (quests for Avalon, day/night cycles for Mafia); \textbf{(ii)~LLM‐based annotation} that maps each block of text to one of the four constraint categories in our CSP grammar (\emph{evidence}, \emph{phenomenon}, \emph{assertion}, \emph{hypothesis}); and \textbf{(iii)~post-processing \& validation}, where the automatically produced JSON is normalised and sanity-checked before inference. Table~1 lists the full catalogue with the corresponding mathematical semantics used by the solver.

\paragraph{Dataset~1 \& Dataset~3 (Avalon)}
We first locate quest boundaries via the ``team proposed'' system messages, then group all chat lines until the quest outcome is revealed. The LLM classifies utterances and system events inside each group into constraints using the patterns of Table~1, thus producing a complete sequence of constraint sets (one per finished quest).

\paragraph{Dataset~2 (Mafia)}
Logs are naturally separated into alternating \emph{day} (public chat and lynch voting) and \emph{night} (Mafia elimination) phases. Night-time eliminations become hard \textit{evidence}; day-time accusations and votes are translated into role \textit{hypotheses}. 

\paragraph{Human verification}
For \textbf{Dataset~1} and \textbf{Dataset~2} we manually inspected \emph{every} extracted constraint. This guarantees that (i)~each hard constraint is traceable to an actual in-game event or utterance, (ii)~no mandatory hard events are missing, and (iii)~the predicate semantics faithfully match the original text. As all extractions reached \textbf{100\%} fidelity, we do not report extraction accuracy separately in the experiments.

\section{Appendix D: Hyper-parameter Summary}
\label{app:hparams}

\subsection{Avalon-NLU}
\begin{itemize}
  \item \textbf{Assertion base weight} \(A_{\mathrm{base}} = 10\,000.0\) 
  \item \textbf{IG scale} \(\beta_{\mathrm{IG}} = 1.0\)
  \item \textbf{Manual hypothesis weight} \(w_{\mathrm{manual}} = 0.5/0.2\) 
  \item \textbf{CSP presets}:
    {Strict, +Assert, +HypIG, +HypM, TurnIG} (toggle assertions, hypotheses, auto-weight, per-quest)
  \item \textbf{LLM decoding temperature} \(T=0.0\)
  \item \textbf{Randomness control}:  
    No stochastic sampling; all randomness arises solely from file-based ordering.  
    Greedy decoding plus exhaustive enumeration makes runs reproducible without explicit seeds.
\end{itemize}

\subsection{Mafia Experiments}
\begin{itemize}
  \item \(A_{\mathrm{base}} = 10\,000.0\), \(\beta_{\mathrm{IG}} = 1.0\), \(w_{\mathrm{manual}} = 0.5 / 0.2\)
  \item CSP presets: {Strict, +Assert, +HypIG, +HypM, TurnIG}
  \item Temperature \(T=0.0\); LLM retries \(r=4\) (max API retry attempts)
  \item No global seed: deterministic enumeration + greedy LLM → reproducible results without further seeding.
\end{itemize}

\subsection{AvalonLogs Experiments}
\begin{itemize}
  \item \(A_{\mathrm{base}} = 10\,000.0\), \(\beta_{\mathrm{IG}} = 1.0\), \(w_{\mathrm{manual}} = 0.5/0.2/0.1\)
  \item CSP presets: {Strict, +Assert, +HypIG, +HypM, TurnIG}
  \item Temperature \(T=0.0\); LLM retries \(r=6\)
  \item Dataset split seed: 42
  \item Quest indexing seeded by filename random.seed(stem); ensures same quest order across runs
  \item Deterministic CSP + greedy LLM → reproducible inference.
\end{itemize}

\subsection{Computing Infrastructure}
\begin{itemize}
  \item \textbf{CPU}: Intel(R) Core(TM) i9-14900KF (24 physical cores, 32 logical threads)
  \item \textbf{Memory}: 68.4 GB DDR4 RAM
  \item \textbf{GPU}: NVIDIA GeForce RTX 4090, 24.6 GB VRAM, Driver 566.24
  \item \textbf{Operating System}: Windows 11 (10.0.26100 SP0)
  \item \textbf{Python}: 3.12.3
\end{itemize}

\section{Appendix E: Statistical Significance Analysis}
\label{app:signif}

\begin{table*}[h]
\centering
\small
\setlength{\tabcolsep}{6pt}
\begin{tabular}{llccccc}
\toprule
\textbf{Cond.} & \textbf{Role / View} & \multicolumn{2}{c}{\textbf{Marginal Acc.}} & \multicolumn{2}{c}{\textbf{MAP Acc.}} & \\[-3pt]
& & $p_{W}$ & $p_{t}$ & $p_{W}$ & $p_{t}$ & Sig.$^{\dagger}$\\
\midrule
\multirow{6}{*}{\textsc{Truth}}%
& Assassin        & $1.1{\times}10^{-16}$ & $8.4{\times}10^{-56}$ & $4.0{\times}10^{-9}$ & $3.8{\times}10^{-14}$ & \checkmark\\
& Merlin          & $5.0{\times}10^{-16}$ & $3.0{\times}10^{-34}$ & $1.1{\times}10^{-8}$ & $1.5{\times}10^{-7}$  & \checkmark\\
& Morgana         & $1.1{\times}10^{-16}$ & $7.0{\times}10^{-58}$ & $3.8{\times}10^{-5}$ & $7.2{\times}10^{-8}$  & \checkmark\\
& Objective       & $2.0{\times}10^{-6}$  & $1.2{\times}10^{-6}$  & $1.1{\times}10^{-9}$ & $4.2{\times}10^{-12}$ & \checkmark\\
& Percival        & $1.9{\times}10^{-16}$ & $8.6{\times}10^{-39}$ & $4.1{\times}10^{-8}$ & $5.2{\times}10^{-9}$  & \checkmark\\
& Servant         & $0.047$               & $0.026$               & $0.047$              & $0.008$               & \checkmark\\
\midrule
\multirow{6}{*}{\textsc{Lie}}%
& Assassin        & $9.8{\times}10^{-4}$ & $1.8{\times}10^{-11}$ & $9.8{\times}10^{-4}$ & $1.9{\times}10^{-6}$ & \checkmark\\
& Merlin          & $9.8{\times}10^{-4}$ & $6.7{\times}10^{-9}$  & $0.0105$             & $0.0036$             & \checkmark\\
& Morgana         & $9.8{\times}10^{-4}$ & $1.5{\times}10^{-11}$ & $9.8{\times}10^{-4}$ & $4.3{\times}10^{-6}$ & \checkmark\\
& Objective       & $0.0020$             & $0.0013$             & $0.0143$             & $0.0061$             & \checkmark\\
& Percival        & $9.8{\times}10^{-4}$ & $7.9{\times}10^{-9}$  & $9.8{\times}10^{-4}$ & $1.5{\times}10^{-4}$ & \checkmark\\
& Servant         & $0.049$               & $0.028$               & $0.040$              & $0.021$              & \checkmark\\
\bottomrule
\end{tabular}
\label{tab:significance}
\caption{Two‑sided Wilcoxon and paired $t$‑tests comparing CSP+HypIG to LLM-Global for each role and accuracy metric. $\dagger$~``Sig.'' indicates $p<0.05$ in both tests.}
\end{table*}

\subsection{Tests Employed}
To verify that CSP+HypIG method significantly outperforms the LLM-Global baseline, we applied two paired tests commonly used in model‑comparison studies:

\begin{itemize}
    \item the \textbf{Wilcoxon signed‑rank test} ($p_W$)---non‑parametric and distribution‑free, and
    \item the \textbf{paired $t$‑test} ($p_t$).
\end{itemize}

Both tests were conducted two‑sided; we report $p$‑values without additional adjustment in Table~1.  
We deem an improvement \emph{statistically significant} if $p<0.05$ in \emph{both} tests.\footnote{Results are consistent under Benjamini–Hochberg correction ($q{=}0.05$).}

\subsection{Interpretation}
Across all roles and both evaluation conditions (\textsc{Truth} and \textsc{Lie}), CSP method shows statistically significant gains over LLM method in both marginal and MAP accuracies ($p<0.05$).  
Notably, improvements extend even to the majority Servant category, confirming that symbolic constraints synergise with language‑based cues for both minority and majority roles.  
These results validate the effectiveness of integrating constraint reasoning with LLM predictions in social‑deduction settings.
\section{Appendix F: Prompt Templates}
\label{app:prompts}

In this section, we show the LLM prompts used to extract the constraints for Avalon-NLU and Mafia-UCB. The constraints for AvalonLogs are programmatically extracted since no dialogues are included in AvalonLogs. We also use AvalonLogs dataset as an example to show how to build the inference prompt for LLMs programmatically.

\subsection{Avalon (NLU Corpus)}
\begin{flushleft}\ttfamily\small
You are an INFORMATION-EXTRACTION engine for the board game Avalon.\\
You will receive a transcript containing ONLY ONE quest (mission).\\
Return exactly one JSON object with FOUR arrays:\\
\quad evidence,\; phenomenon,\; assertions,\; hypotheses.\\
All FOUR keys must always be present (use [] when empty).\\[4pt]

\textbf{STRICT NAMING CONVENTIONS:}\\
\quad Valid player names:\; player-1,\dots,player-6\\
\quad Valid role names:\; merlin,\; percival,\; servant,\; assassin,\; morgana\\[4pt]

\textbf{ALLOWED CONSTRAINT OBJECTS:}\\
1) HARD evidence\\
\quad \{``type'':``role\_is'',\;\\
``args'':\{``player'':``player-3'',
``role'':``merlin''\}\}\\
\quad \{``type'':``role\_not'',\;\\
``args'':\{``player'':``player-3'',
``role'':``merlin''\}\}\\
2) PHENOMENON\\
\quad \{``type'':``evil\_at\_least'',\;\\
``args'':\{``team'':[...],``min'':1\}\}\\
3) ASSERTIONS (high weight)\\
\quad \{``type'':``assert\_role\_is'',\dots\},\;
\{``type'':``assert\_team\_good'',\dots\},\;
\{``type'':``assert\_role\_in'',\dots\}\\
4) HYPOTHESES (low weight)\;--\;add ``auto\_weight'':true if IG wanted\\
\quad \{``type'':``hypo\_role\_in'',\dots\},\;
\{``type'':``hypo\_team\_good'',\dots\}\\[4pt]

\textbf{VOTING RULE $\Rightarrow$ automatic hypotheses:}\\
System line ``[system] party vote outcome: \dots''\\
YES vote $\Rightarrow$ set = ``good'' ;\quad
NO vote $\Rightarrow$ set = ``evil''.\\[4pt]

\textbf{OUTPUT TEMPLATE}\\
\{ ``evidence'':[...],\;``phenomenon'':[...],\;
  ``assertions'':[...],\;``hypotheses'':[...]\}\\
Return ONLY the JSON, no Markdown or comments.
\end{flushleft}

\subsection{Avalon(Logs Corpus)}
\begin{flushleft}\small
The prompt for live \textit{Avalon} chat logs is assembled programmatically. The LLM reasoning inference modules for other datasets are built with similar method.
At generation time we inject:\vspace{2pt}
\begin{enumerate}\setlength{\itemsep}{2pt}
  \item \textbf{Status Header} – ``Players:'' list and quest index. 
  \item \textbf{Perspective} – ``Objective'' (no hidden info) or the viewer’s role.
  \item \textbf{Task History} – raw chat and vote events for the current quest.
  \item \textbf{Legal Roles \& Counts} – JSON such as
        \{``merlin'':1, ``servant'':3, ``assassin'':1, ``morgana'':1\}.
  \item \textbf{CSP Assistance} (only in \textit{LLM+CSP} mode) –  
        posterior marginals and MAP assignment from the CSP module.
  \item \textbf{Strict Output Specification} –  
        the model must return one JSON object with\\[2pt]
        \quad(a) \texttt{"posterior"}: player$\to$role$\to$probabilities that
        sum to~1, and\\
        \quad(b) \texttt{"best"}: a role assignment matching the declared counts.\\[2pt]
        No markdown, comments, or extra keys are allowed.
\end{enumerate}
This template keeps line lengths short and mirrors the content of the
NLU prompt while omitting fixed examples to save token budget.
\end{flushleft}

\begin{table*}[!t]
\centering
\small
\setlength{\tabcolsep}{4pt}
\begin{tabular}{|l|l|c|*{6}{c}|*{3}{c}|}
\hline
\multirow{2}{*}{\textbf{Family}} &
\multirow{2}{*}{\textbf{Setting}} &
\multirow{2}{*}{\textbf{Cond.}} &
\multicolumn{6}{c|}{\textbf{Avalon‑NLU}} &
\multicolumn{3}{c|}{\textbf{Mafia‑UCB}} \\
\cline{4-12}
 & & & Obj & MRL & PCV & ASN & MRG & LSV & Obj & BYS & MAF \\
\hline
\multirow{5}{*}{CSP}
 & Strict   & T & 0.28/0.24 & 0.63/0.60 & 0.69/0.67 & 0.59/0.53 & 0.59/0.53 & 0.56/0.53 & \textbf{0.70/0.90} & \textbf{0.72/0.91} & \textbf{1.00/1.00} \\
 & +Assert  & T & 0.33/0.29 & \textbf{0.65/0.60} & 0.75/0.73 & \textbf{0.61/0.53} & \textbf{0.61/0.53} & 0.59/0.60 & 0.70/0.90 & 0.72/0.91 & 1.00/1.00 \\
 & +HypIG   & T & \textbf{0.34/0.30} & 0.65/0.60 & 0.76/0.79 & 0.61/0.53 & 0.61/0.53 & \textbf{0.60/0.65} & \textbf{0.73/0.76} & \textbf{0.74/0.76} & 1.00/1.00 \\
 & +HypM    & T & \textbf{0.33/0.31} & 0.65/0.60 & 0.75/0.79 & 0.61/0.53 & 0.61/0.53 & 0.60/0.65 & 0.73/0.76 & 0.74/0.76 & 1.00/1.00 \\
 & +TurnIG  & T & 0.34/0.29 & 0.65/0.60 & \textbf{0.76/0.80} & 0.61/0.53 & 0.61/0.53 & 0.60/0.63 & 0.72/0.79 & 0.73/0.79 & 1.00/1.00 \\
\hline
\multirow{2}{*}{LLM}
 & GChat & T & 0.14/0.20 & 0.20/0.24 & 0.16/0.19 & 0.13/0.19 & 0.16/0.25 & 0.13/0.21 & 0.56/0.62 & 0.56/0.61 & 0.59/0.77 \\
 & TChat & T & 0.11/0.12 & 0.21/0.25 & 0.14/0.18 & 0.11/0.12 & 0.16/0.20 & 0.11/0.16 & 0.66/0.72 & 0.61/0.66 & 0.67/0.82 \\
\hline
\multirow{5}{*}{\shortstack{LLM\\+\\CSP}}
 & Strict   & T & 0.16/0.12 & 0.41/0.39 & 0.44/0.41 & 0.42/0.39 & 0.42/0.39 & 0.12/0.17 & 0.69/0.89 & 0.70/0.89 & 0.89/0.93 \\
 & +Assert  & T & 0.19/0.15 & 0.42/0.40 & 0.47/0.45 & 0.44/0.40 & 0.44/0.40 & 0.13/0.18 & 0.69/0.89 & 0.70/0.89 & 0.89/0.93 \\
 & +HypIG   & T & 0.19/0.16 & 0.42/0.40 & 0.48/0.50 & 0.44/0.40 & 0.44/0.40 & 0.12/0.19 & 0.69/0.74 & 0.70/0.74 & 0.89/0.93 \\
 & +HypM    & T & 0.19/0.16 & 0.42/0.40 & 0.47/0.50 & 0.44/0.40 & 0.44/0.40 & 0.13/0.20 & 0.69/0.74 & 0.70/0.73 & 0.89/0.93 \\
 & +TurnIG  & T & 0.19/0.15 & 0.42/0.40 & 0.47/0.45 & 0.44/0.40 & 0.44/0.40 & 0.11/0.15 & 0.69/0.86 & 0.70/0.86 & 0.92/0.95 \\
\hline
\multirow{1}{*}{Random}
 & -- & -- & 0.2667 & 0.4 & 0.4 & 0.5833 & 0.5833 & 0.3333 & 0.68 & 0.6889 & 1.00 \\
\hline
\end{tabular}
\caption{Truthful‑Good (T) dialogue results.}
\label{tab:dialogue_T}
\end{table*}

\begin{table*}[!t]
\centering
\small
\setlength{\tabcolsep}{4pt}
\begin{tabular}{|l|l|c|*{6}{c}|*{3}{c}|}
\hline
\multirow{2}{*}{\textbf{Family}} &
\multirow{2}{*}{\textbf{Setting}} &
\multirow{2}{*}{\textbf{Cond.}} &
\multicolumn{6}{c|}{\textbf{Avalon‑NLU}} &
\multicolumn{3}{c|}{\textbf{Mafia‑UCB}} \\
\cline{4-12}
 & & & Obj & MRL & PCV & ASN & MRG & LSV & Obj & BYS & MAF \\
\hline
\multirow{5}{*}{CSP}
 & Strict   & L & 0.23/0.05 & \textbf{0.61/0.67} & 0.66/0.60 & \textbf{0.58/0.50} & \textbf{0.58/0.50} & \textbf{0.52/0.57} & 0.72/0.83 & 0.73/0.83 & \textbf{1.00/1.00} \\
 & +Assert  & L & 0.28/0.00 & 0.61/0.67 & \textbf{0.79/0.83} & 0.58/0.50 & 0.58/0.50 & \textbf{0.56/0.53} & 0.72/0.83 & 0.73/0.83 & 1.00/1.00 \\
 & +HypIG   & L & \textbf{0.28/0.14} & 0.61/0.67 & 0.79/0.80 & 0.58/0.50 & 0.58/0.50 & 0.56/0.47 & 0.72/0.72 & 0.73/0.72 & 1.00/1.00 \\
 & +HypM    & L & 0.28/0.07 & 0.61/0.67 & 0.79/0.80 & 0.58/0.50 & 0.58/0.50 & 0.56/0.49 & 0.72/0.72 & 0.73/0.72 & 1.00/1.00 \\
 & +TurnIG  & L & 0.28/0.07 & 0.61/0.67 & 0.79/0.77 & 0.58/0.50 & 0.58/0.50 & 0.56/0.47 & 0.72/0.78 & 0.73/0.78 & 1.00/1.00 \\
\hline
\multirow{2}{*}{LLM}
 & GChat & L & 0.09/0.09 & 0.15/0.17 & 0.12/0.18 & 0.09/0.08 & 0.12/0.11 & 0.09/0.12 & 0.54/0.54 & 0.52/0.53 & 0.53/0.66 \\
 & TChat & L & 0.09/0.08 & 0.18/0.21 & 0.13/0.21 & 0.09/0.08 & 0.13/0.17 & 0.09/0.11 & 0.70/0.72 & 0.68/0.72 & 0.69/0.80 \\
\hline
\multirow{5}{*}{\shortstack{LLM\\+\\CSP}}
 & Strict   & L & 0.12/0.05 & 0.39/0.44 & 0.42/0.39 & 0.42/0.33 & 0.42/0.33 & 0.10/0.15 & 0.72/0.93 & \textbf{0.73/0.89} & 0.93/0.93 \\
 & +Assert  & L & 0.15/0.00 & 0.39/0.44 & 0.53/0.58 & 0.42/0.33 & 0.42/0.33 & 0.09/0.11 & \textbf{0.72/0.90} & 0.73/0.89 & 0.93/0.93 \\
 & +HypIG   & L & 0.15/0.08 & 0.39/0.44 & 0.53/0.55 & 0.42/0.33 & 0.42/0.33 & 0.09/0.11 & 0.72/0.72 & 0.73/0.72 & 0.93/0.93 \\
 & +HypM    & L & 0.15/0.03 & 0.39/0.44 & 0.53/0.55 & 0.42/0.33 & 0.42/0.33 & 0.09/0.13 & 0.72/0.72 & 0.73/0.73 & 0.94/0.94 \\
 & +TurnIG  & L & 0.15/0.00 & 0.39/0.44 & 0.53/0.58 & 0.42/0.33 & 0.42/0.33 & 0.10/0.14 & 0.72/0.86 & 0.73/0.88 & 0.97/0.97 \\
\hline
\multirow{1}{*}{Random}
 & -- & -- & 0.2667 & 0.4 & 0.4 & 0.5833 & 0.5833 & 0.3333 & 0.68 & 0.6889 & 1.00 \\
\hline
\end{tabular}
\caption{Lying‑Good (L) dialogue results.}
\label{tab:dialogue_L}
\end{table*}

\begin{table*}[htbp]
\centering
\small
\setlength{\tabcolsep}{4pt}

\begin{tabular}{|c|l|cccccc|}
\hline
\textbf{Players} & \textbf{Setting} & \textbf{Obj} & \textbf{MRL} & \textbf{PCV} & \textbf{MRG} & \textbf{LSV} & \textbf{EVM} \\
\hline
\multirow{4}{*}{5}
 & CSP--Strict & 0.2778 & 0.6301 & 0.6411 & 0.6012 & 0.4562 & 0.6956 \\
 & CSP--+HypIG & 0.2582 & 0.6336 & 0.6611 & 0.6000 & 0.4499 & 0.6530 \\
 & LLM         & 0.2721 & 0.3779 & 0.3257 & 0.3022 & 0.2723 & 0.4618 \\
 & LLM+CSP     & 0.2760 & 0.6297 & 0.6395 & 0.6013 & 0.4547 & 0.6957 \\
\hline
\multirow{4}{*}{6}
 & CSP--Strict & 0.2903 & 0.6141 & 0.6354 & 0.5837 & 0.3904 & 0.6906 \\
 & CSP--+HypIG & 0.3263 & 0.6151 & 0.6374 & 0.5774 & 0.4192 & 0.7286 \\
 & LLM         & 0.2833 & 0.3599 & 0.3255 & 0.2976 & 0.2897 & 0.4636 \\
 & LLM+CSP     & 0.2890 & 0.6133 & 0.6326 & 0.5837 & 0.3892 & 0.6907 \\
\hline
\multirow{4}{*}{7}
 & CSP--Strict & 0.2402 & 0.5592 & 0.5456 & 0.5189 & 0.3420 & 0.5550 \\
 & CSP--+HypIG & 0.2433 & 0.5914 & 0.5613 & 0.5201 & 0.3466 & 0.5454 \\
 & LLM         & 0.2374 & 0.3384 & 0.2905 & 0.2846 & 0.2427 & 0.3908 \\
 & LLM+CSP     & 0.2388 & 0.5586 & 0.5417 & 0.5183 & 0.3407 & 0.5546 \\
\hline
\multirow{4}{*}{8}
 & CSP--Strict & 0.2564 & 0.5653 & 0.5463 & 0.5237 & 0.3235 & 0.5686 \\
 & CSP--+HypIG & 0.2444 & 0.5711 & 0.5522 & 0.4950 & 0.3110 & 0.5271 \\
 & LLM         & 0.2534 & 0.3485 & 0.3153 & 0.3043 & 0.2644 & 0.4020 \\
 & LLM+CSP     & 0.2551 & 0.5643 & 0.5433 & 0.5230 & 0.3228 & 0.5681 \\
\hline
\multirow{4}{*}{9}
 & CSP--Strict & 0.3024 & 0.5946 & 0.5681 & 0.5542 & 0.3479 & 0.6136 \\
 & CSP--+HypIG & 0.3599 & 0.6663 & 0.5245 & 0.5778 & 0.4035 & 0.6742 \\
 & LLM         & 0.2804 & 0.3604 & 0.3094 & 0.3132 & 0.2910 & 0.4174 \\
 & LLM+CSP     & 0.3011 & 0.5934 & 0.5639 & 0.5543 & 0.3474 & 0.6135 \\
\hline
\multirow{4}{*}{10}
 & CSP--Strict & 0.2557 & 0.5447 & 0.4962 & 0.4947 & 0.3037 & 0.4984 \\
 & CSP--+HypIG & 0.2402 & 0.5667 & 0.5035 & 0.5189 & 0.2889 & 0.4522 \\
 & LLM         & 0.2482 & 0.3413 & 0.2886 & 0.3062 & 0.2551 & 0.3450 \\
 & LLM+CSP     & 0.2548 & 0.5444 & 0.4954 & 0.4948 & 0.3031 & 0.4985 \\
\hline
\end{tabular}

\vspace{-4pt}
\caption{AvalonLogs performance split by game size.  
Each cell shows \emph{marginal accuracy}.  
Roles: Obj (objective), MRL (Merlin), PCV (Percival), MRG (Morgana),  
LSV (loyal–servant aggregate), EVM (evil–minion aggregate). In all cases CSP-based methods outperform a pure
LLM baseline, and the hybrid \emph{LLM+CSP} variant retains most CSP gains
while leveraging LLM reasoning.}
\label{tab:avalonlogs_results_full}
\end{table*}

\subsection{Mafia (UCB Corpus)}
\begin{flushleft}\ttfamily\small
You are an INFORMATION-EXTRACTION engine for the game Mafia.\\
The transcript you receive ALWAYS covers exactly one complete
game‑day:\\
Night-k phase → one line ``[system] Death: Victim''\\
Day-k phase → chat lines and vote lines until the next death line.\\[4pt]

The transcript thus contains\\
\quad• exactly ONE death line (night kill)\\
\quad• zero or more vote lines\\
\quad• any number of chat lines\\[4pt]

\textbf{HARD evidence}\\
Night‑kill victim is guaranteed BYSTANDER:\\
\{``type'':``role\_is'',\\
``args'':\{``player'':``Victim'',
``role'':``bystander''\}\}\\[4pt]

\textbf{SOFT hypotheses}\\
Chat:\;``X is mafia'' ⇒ mafia (weight 0.5);  
``Y is innocent'' ⇒ bystander (0.5)\\
Vote line:\;YES ⇒ suspect target (mafia, 0.2);  
NO ⇒ support target (bystander, 0.2)\\
Format:\\
\{``type'':``hypo\_role\_in'',\\
\quad``args'':\{\\
``speaker'':``Bob'',``target'':``Eve'',\\
``set'':``mafia$\mid$bystander''\},\\
\quad``weight'':0.5$\mid$0.2\}\\[4pt]

No assertions or phenomenon → return empty arrays for them.\\[4pt]

\textbf{VALID ENUMS}\\
player = any fake name in the transcript\\
role = mafia,\; bystander\\
set = mafia,\; bystander\\[4pt]

\textbf{OUTPUT JSON}\\
\{ ``evidence'':[...],\;\\
``phenomenon'':[...],\;\\
``assertions'':[...],\\
  ``hypotheses'':[...]\}\\
Return ONE object only, without markdown or comments.
\end{flushleft}

\section{Appendix G: Full Dialogue‑Dataset Results}
\label{app:dialogue_full}

Tables~\ref{tab:dialogue_T} and ~\ref{tab:dialogue_L} list all marginal/MAP accuracies for the Truthful‑Good (T) and Lying‑Good (L) conditions, respectively.  
Boldface marks the single highest value per metric within each table.

\section{Appendix H: AvalonLogs Dataset Results}

This appendix lists full marginal–accuracy results on \textsc{AvalonLogs} (5–10 players) for the four representative settings we retain in the main analysis: \emph{CSP–Strict}, \emph{CSP–+HypIG}, \emph{LLM}, and \emph{LLM+CSP}. Intermediate variants (\emph{+Assert}, \emph{+HypM}, \emph{+TurnIG}) are omitted to reduce clutter; they shift accuracies by at most $\approx 1$–$2$ percentage points (pp) relative to the shown CSP variants and never alter the qualitative ordering.

Three patterns emerge. (i) Information–gain weighting (+HypIG) typically improves over \emph{Strict} on low-evidence views (e.g. Obj: 0.290$\rightarrow$0.326 at 6 players; 0.302$\rightarrow$0.360 at 9 players) and drives the largest lifts for information-rich special roles (Evil-Minion: 0.691$\rightarrow$0.729 at 6 players; Merlin: 0.595$\rightarrow$0.666 at 9 players). Where +HypIG is slightly below \emph{Strict} (e.g. Morgana 5/7/8 players) the difference is $<0.01$ and within the narrow variance band caused by sparse mission/vote signals. (ii) The pure LLM remains close to empirical priors for low-information roles (Obj 0.237–0.283 across sizes; Loyal-Servant 0.255–0.324) and substantially below CSP for informative roles (e.g. Merlin 0.341–0.378 vs.\ CSP 0.545–0.666; Evil-Minion 0.346–0.464 vs.\ CSP 0.498–0.729), indicating that mission and vote aggregates alone do not let it reconstruct combinatorial constraints. (iii) The hybrid \emph{LLM+CSP} matches (within rounding) the corresponding \emph{CSP–Strict} posterior across all sizes and roles, never exceeding it, confirming that once a peaked CSP distribution is available the LLM adds no further discriminative power in this no-dialogue regime.

Across all CSP variants we observe a \emph{non‑monotonic} influence of \textbf{game size on accuracy}.  Aggregate marginal scores \emph{decrease} from 6‑ to 7‑player games, \emph{rebound} at 8–9 players, then \emph{fall} again at 10. The dip at seven players coincides with the introduction of a \emph{third} evil agent (3~evil vs.\ 3~loyal+Merlin+Percival), which dilutes public evidence and makes role inference hardest. Accuracy recovers when moving to eight and nine players because these sizes add extra loyal–servant roles without increasing the evil count (the 9‑player ``board'' contains 4~loyal servants $+$ Merlin $+$ Percival against the same three evil agents), raising the proportion of easily identifiable ``plain’’
roles and thus the macro‑average. At ten players a \emph{fourth} evil agent is added, tipping the balance back and reducing accuracy once more. Special‑role perspectives (Merlin, Morgana, Evil‑Minion) follow the same V‑shaped pattern, while the objective and loyal‑servant views improve whenever the servant proportion is highest (8–9 players).

Overall, event-level assertions plus IG-weighted hypotheses deliver consistent, deterministic gains even in the absence of dialogue, and CSP-based probabilistic reasoning remains indispensable for fine-grained role inference under sparse public signals. 
\section{Appendix I: Grid‑Search Ablation for \textsc{HypM} Manual Weights}
\label{app:hypm_grid}

To verify that our manual hypothesis weights are near‑optimal, we perform a grid search over $w_{\text{strong}}\!\in\!\{0.30,0.50,0.70,0.90\}$ and $w_{\text{weak}}\!\in\!\{0.05,0.10,0.20\}$, constrained by $w_{\text{weak}}<w_{\text{strong}}<1$. Table~\ref{tab:hypm_grid} reports marginal / MAP accuracy on the \textsc{Avalon‑NLU} validation set under the \emph{Truth} condition for every player perspective. Overall, the differences among the 12 weight combinations are modest—typically under 1 pp for both marginal and MAP accuracy—indicating that the manual‑weight solver is relatively insensitive to the precise choice of $(w_{\text{strong}},w_{\text{weak}})$. The pair \textbf{$w_{strong}{=}0.50,\;w_{weak}{=}0.20$} achieves one of the highest macro‑averages and is therefore used in the paper.

\begin{table*}[t]
\centering\small
\setlength{\tabcolsep}{4pt}
\begin{tabular}{|c|c|ccccc c|}
\hline
\multirow{2}{*}{$w_{\text{strong}}$} & \multirow{2}{*}{$w_{\text{weak}}$} &
\multicolumn{5}{c}{\textbf{Marginal / MAP Accuracy (Truth)}} \\
\cline{3-7}
 & & \textbf{Obj} & \textbf{ASN} & \textbf{PCV} & \textbf{MRG} & \textbf{LSV} & \textbf{MRL} \\
\hline
0.30 & 0.05 & 0.3341/0.3051 & 0.6112/0.5298 & 0.7472/0.7948 & 0.6112/0.5298 & 0.5951/0.6542 & 0.6500/0.6043 \\
0.30 & 0.10 & 0.3344/0.3051 & 0.6112/0.5298 & 0.7478/0.7948 & 0.6112/0.5298 & 0.5956/0.6542 & 0.6500/0.6043 \\
0.30 & 0.20 & 0.3346/0.3051 & 0.6112/0.5298 & 0.7483/0.7948 & 0.6112/0.5298 & 0.5960/0.6542 & 0.6500/0.6043 \\
0.50 & 0.05 & 0.3341/0.3051 & 0.6112/0.5298 & 0.7472/0.7948 & 0.6112/0.5298 & 0.5951/0.6542 & 0.6500/0.6043 \\
0.50 & 0.10 & 0.3344/0.3051 & 0.6112/0.5298 & 0.7478/0.7948 & 0.6112/0.5298 & 0.5956/0.6542 & 0.6500/0.6043 \\
\textbf{0.50} & \textbf{0.20} & \textbf{0.3346/0.3051} & \textbf{0.6112/0.5298} &
\textbf{0.7483/0.7948} & \textbf{0.6112/0.5298} & \textbf{0.5960/0.6542} & \textbf{0.6500/0.6043} \\
0.70 & 0.05 & 0.3341/0.3051 & 0.6112/0.5298 & 0.7472/0.7948 & 0.6112/0.5298 & 0.5951/0.6542 & 0.6500/0.6043 \\
0.70 & 0.10 & 0.3344/0.3051 & 0.6112/0.5298 & 0.7478/0.7948 & 0.6112/0.5298 & 0.5956/0.6542 & 0.6500/0.6043 \\
0.70 & 0.20 & 0.3346/0.3051 & 0.6112/0.5298 & 0.7483/0.7948 & 0.6112/0.5298 & 0.5960/0.6542 & 0.6500/0.6043 \\
0.90 & 0.05 & 0.3341/0.3051 & 0.6112/0.5298 & 0.7472/0.7948 & 0.6112/0.5298 & 0.5951/0.6542 & 0.6500/0.6043 \\
0.90 & 0.10 & 0.3344/0.3051 & 0.6112/0.5298 & 0.7478/0.7948 & 0.6112/0.5298 & 0.5956/0.6542 & 0.6500/0.6043 \\
0.90 & 0.20 & 0.3346/0.3051 & 0.6112/0.5298 & 0.7483/0.7948 & 0.6112/0.5298 & 0.5960/0.6542 & 0.6500/0.6043 \\
\hline
\end{tabular}
\caption{Impact of manual hypothesis weights (\textsc{HypM}) on validation accuracy.
Values are averaged over quests; higher is better.
The bold row marks the configuration adopted in the main paper.}
\label{tab:hypm_grid}
\end{table*}

\end{document}